\documentclass[preprint,12pt]{elsarticle}

\usepackage{amssymb}
\usepackage{amsmath}
\usepackage{bm}
\usepackage{subfig}
\usepackage{wrapfig}
\usepackage{lipsum}     
\usepackage{float} 
\usepackage{graphicx}           
\usepackage{url} 
\usepackage{soul}
\usepackage{enumitem}
\usepackage{lineno}
\usepackage{diagbox}
\usepackage{color}

\journal{Computational Mechanics}

\begin{document}

\begin{frontmatter}


\title{Conditional deep surrogate models for stochastic, high-dimensional, and multi-fidelity systems}



\author{Yibo Yang}
\author{Paris Perdikaris}

\address{Department of Mechanical Engineering and Applied Mechanics, \\
University of Pennsylvania, \\
Philadelphia, PA 19104, USA}

\begin{abstract}
We present a probabilistic deep learning methodology that enables the construction of predictive data-driven surrogates for stochastic systems. Leveraging recent advances in variational inference with implicit distributions, we put forth a statistical inference framework that enables the end-to-end training of surrogate models on paired input-output observations that may be stochastic in nature, originate from different information sources of variable fidelity, or be corrupted by complex noise processes. The resulting surrogates can accommodate high-dimensional inputs and outputs and are able to return predictions with quantified uncertainty. The effectiveness our approach is demonstrated through a series of canonical studies, including the regression of noisy data, multi-fidelity modeling of stochastic processes, and uncertainty propagation in high-dimensional dynamical systems.
\end{abstract}

\begin{keyword}
Probabilistic deep learning \sep Generative adversarial networks \sep Variational inference \sep Multi-fidelity modeling \sep Data-driven surrogates


\end{keyword}

\end{frontmatter}


\section{Introduction}\label{sec:Introduction}

The analysis of complex systems can often be significantly expedited through the use of surrogate models that aim to minimize the need of repeatedly evaluating the true data-generating process, let that be a costly experimental assay or a large-scale computational model. The task of building surrogate models in essence defines a {\em supervised learning} problem in which one aims to distill the governing relation between system inputs and outputs. A successful completion of this task yields a simple and cheap mechanism for predicting the system's response for a new, previously unobserved input, which can be subsequently used to accelerate downstream tasks such as optimization loops, uncertainty quantification, and sensitivity analysis studies. 

Fueled by recent developments in data analytics and machine learning, data-driven approaches to building surrogate models have been gaining great popularity among diverse scientific disciplines. We now have a collection of techniques that have enabled progress across a wide spectrum of applications, including  design optimization \cite{forrester2007multi, robinson2008surrogate, alexandrov2001approximation}, the design of materials \cite{sun2010two, sun2011multi} and supply chains \cite{celik2010dddas}, model calibration \cite{perdikaris2016model, perdikaris2015data, perdikaris2015calibration}, and uncertainty quantification \cite{eldred2009comparison, ng2012multifidelity,padron2014multi, biehler2015towards, peherstorfer2016optimal, peherstorfer2016multifidelity, peherstorfer2016survey, narayan2014stochastic, zhu2014computational, bilionis2013multi, parussini2017multi,perdikaris2016multifidelity}. 
Such approaches are built on the premise of treating the true data-generating process as a black-box, and try to construct parametric surrogates of some form $\bm{y}=f_{\theta}(\bm{x})$ directly from observed input-output pairs $\{\bm{x},\bm{y}\}$. Except perhaps for Gaussian process regression models \cite{rasmussen2004gaussian} that rely on a probabilistic formulation for quantifying predictive uncertainty, most existing approaches use theoretical error bounds to assess the accuracy of the surrogate model predictions and are formulated based on rather limiting assumptions on the form $f$ (e.g., linear or smooth nonlinear). Despite the growing popularity of data-driven surrogate models, a key challenge that still remains open pertains to cases where the entries in $\bm{x}$ and $\bm{y}$ are high-dimensional objects with multiple modalities: vectors with hundreds/thousands of entries, images with thousands/millions of pixels, graphs with thousands of nodes, or even continuous functions and vector fields. Even less well understood is how to build surrogate models for stochastic systems, and how to retain predictive robustness in cases where the observed data is corrupted by complex noise processes.

In this work we aim to formulate, implement, and study novel probabilistic surrogate models models in the context of probabilistic data fusion and multi-fidelity modeling of stochastic systems. Unlike existing approaches to surrogate and  multi-fidelity modeling, the proposed methods scale well with respect to the dimension of the input and output data, as well as the total number of training data points. The resulting generative models provide enhanced capabilities in learning arbitrarily complex conditional distributions and cross-correlations between different data sources, and can accommodate data that may be corrupted by correlated and non-Gaussian noise processes. To achieve these goals, we put forth a regularized adversarial inference framework that goes beyond Gaussian and mean field approximations, and has the capacity to seamlessly model complex statistical and functional dependencies in the data, remain robust with respect to non-Gaussian measurement noise, discover nonlinear low-dimensional embeddings through the use of latent variables, and is applicable across a wide range of supervised tasks.

This paper is structured as follows. In section \ref{sec:VI} we provide a comprehensive description of conditional generative models and recent advances in variational inference that have enabled their scalable training. In section \ref{sec:VAE}, we review recent findings that pinpoint the limitations of mean-field variational inference approaches and motivate the use of implicit parametrizations and adversarial learning schemes. Sections \ref{sec:density_ratio}-\ref{sec:predictions} provide a comprehensive discussion on how such schemes can be trained on paired input-output observations $\{\bm{x},\bm{y}\}$ to yield effective approximations of the conditional density $p(\bm{y}|\bm{x})$. In section \ref{sec:results} we will test effectiveness our approach on a series of canonical studies, including the regression of noisy data, multi-fidelity modeling of stochastic processes, and uncertainty propagation in high-dimensional dynamical systems. Finally, section \ref{sec:discussion} summarizes our concluding remarks, while in   \ref{sec:appendix} we provide a comprehensive collection of systematic studies that aim to elucidate the robustness of the proposed algorithms with respect to different parameter settings. All data and code accompanying this manuscript will be made available at \url{https://github.com/PredictiveIntelligenceLab/CADGMs}.

\section{Methods}\label{sec:Methods}

The focal point of this work is formulating, learning, and exploiting general probabilistic models of the form $p(\bm{y}|\bm{x})$.
One one hand, conditional probability models $p(\bm{y}|\bm{x})$ aim to capture the statistical dependence between realizations of deterministic or stochastic input/output pairs $\{\bm{x},\bm{y}\}$, and encapsulate a broad class of problems generally referred to as {\em supervised learning} problems. Take for example the setting in which we would like to characterize the  properties of a material using molecular dynamics simulations. There, $\bm{x}\in\mathbb{R}^{3N}$ corresponds to a thermodynamically valid configuration of all the $N$ particles in the system, $p(\bm{x})$ is the Boltzmann distribution, and $\bm{y}\in\mathbb{R}^{M}$ is a collection of $M$ correlated quantities of interest that characterize the macroscopic behavior of the system (e.g., Young's modulus, ion mobility, optical spectrum properties, etc.). Given some realizations $\{\bm{x}_s,\bm{y}_s\}$, $s=1,\dots,S$, our goal is to {\em learn} a conditional probability model $p(\bm{y}|\bm{x})$ that not only allows us to accurately predict $\bm{y}^{\ast}$ for a new $\bm{x}^{\ast}$ (e.g., by estimating the expectation $\mathbb{E}_{p}(\bm{y}^{\ast}|\bm{x},\bm{y}, \bm{x}^{\ast})$), but, more importantly, it characterizes the complete statistical dependence of $\bm{y}$ on $\bm{x}$, thus allowing us to quantify the uncertainty associated with our predictions. As $N$ is typically very large, this defines a challenging high-dimensional regression problem. Coming to our rescue, is our ability to extract a meaningful and robust representations of the original data that exploits its structure through the use of latent variables. 

\subsection{Variational inference for conditional deep generative models}\label{sec:VI}

\begin{figure}
\centering
\includegraphics[width=\textwidth]{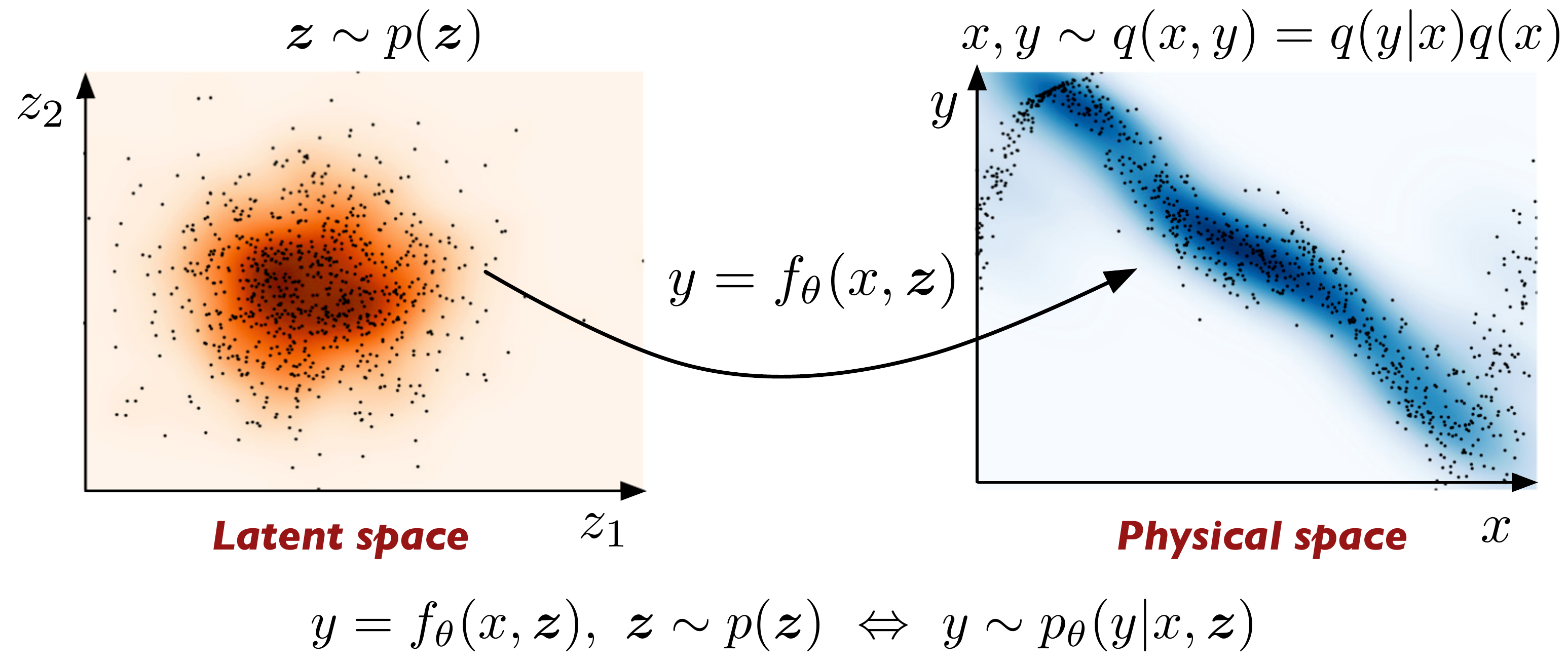}
\caption{{\em Building probabilistic surrogates using conditional generative models:} We assume that each observed data pair in the physical space $(x,y)$ is generated by a deterministic nonlinear transformation of the inputs $x$ and a set of latent variables $\bm{z}$, i.e. $y = f_{\theta}(x,\bm{z})$. This construction generalizes the classical observation model used in regression, namely $y = f_{\theta}(x) + \epsilon$, which can be viewed as a simplified case corresponding to an additive noise model.}
\label{fig:CDGM}
\end{figure}

The introduction of latent variables allows us to express $p(\bm{y}|\bm{x})$ as an infinite mixture model,

\begin{equation}
\label{eq:PSC}
p(\bm{y}|\bm{x}) = \int p(\bm{y},\bm{z}|\bm{x}) d\bm{z} = \int p(\bm{y}|\bm{x},\bm{z})p(\bm{z}|\bm{x}) d\bm{z},
\end{equation}
where $p(\bm{z}|\bm{x})$ is a prior distribution on the latent variables. 
Essentially, this construction postulates that every output $\bm{y}$ in the observed physical space is generated by a transformation of the inputs $\bm{x}$ and a set of latent variables $\bm{z}$, , i.e. $y = f_{\theta}(x,\bm{z})$, where $f_{\theta}$ is a parametrized nonlinear transformation (see figure \ref{fig:CDGM}).This construction generalizes the classical observation model used in regression, namely $y = f_{\theta}(x) + \epsilon$, which can be viewed as a simplified case corresponding to an additive noise model.

Equation \ref{eq:PSC} resembles a mixture model as for every possible value of $\bm{z}$, we add another conditional distribution to $p(\bm{y}|\bm{x})$, weighted by its probability. Now, it is interesting to ask what the latent variables $\bm{z}$ are, given an input/output pair $\{\bm{x},\bm{y}\}$. Namely, from a Bayesian standpoint, we would like to know the posterior distribution $p(\bm{z}|\bm{x},\bm{y})$. However, in general, the relationship between $\bm{z}$ and $\{\bm{x},\bm{y}\}$ can be highly non-linear and both the dimensionality of our observations $\{\bm{x},\bm{y}\}$, and the dimensionality of the latent variables $\bm{z}$, can be quite large. Since both marginal and posterior probability distributions require evaluation of the integral in equation \ref{eq:PSC}, they are intractable.

The seminal work of Kingma and Welling \cite{kingma2013auto} introduced an effective framework for approximating the true underlying conditional $p(\bm{y}|\bm{x})$ with a parametrized distribution $p_{\theta}(\bm{y}|\bm{x})$ that depends on a set of parameters $\theta$. Specifically, they  introduced a parametrized approximating distribution $q_{\phi}(\bm{z}|\bm{x},\bm{y})$ to approximate the true intractable posterior $p(\bm{z}|\bm{x},\bm{y})$, and derived a computable variational objective for estimating the model parameters $\{\theta, \phi\}$ using stochastic optimization \cite{kingma2013auto}. This objective, often referred to as the evidence lower bound (ELBO), provides a tractable lower bound to the marginal likelihood of the model, and takes the form \cite{sohn2015learning}

\begin{equation}
\label{eq:ELBO}
-\log p_{\theta}(\bm{y}|\bm{x}) \le \mathbb{KL}\left[q_{\phi}(\bm{z}|\bm{x},\bm{y})||p(\bm{z}|\bm{x})\right] - 
\mathbb{E}_{\bm{z}\sim q_{\phi}(\bm{z}|\bm{x},\bm{y})} \left[\log p_{\theta}(\bm{y}|\bm{x},\bm{z})\right],
\end{equation}
where $\mathbb{KL}\left[q_{\phi}(\bm{z}|\bm{x},\bm{y})||p(\bm{z}|\bm{x})\right]$ denotes the Kullback-Leibler divergence between the approximate posterior $q_{\phi}(\bm{z}|\bm{x},\bm{y})$ and the prior over the latent variables $p(\bm{z}|\bm{x})$ \cite{kingma2013auto,sohn2015learning}. Due to the resemblance of this approach to neural network auto-encoders \cite{vincent2008extracting,vincent2010stacked}, the model proposed by Kingma and Welling has been coined as the variational auto-encoder, and  the resulting approximate distributions $q_{\phi}(\bm{z}|\bm{x},\bm{y})$ and  $p_{\theta}(\bm{y}|\bm{x},\bm{z})$ are usually referred to as the {\it encoder} and {\it decoder} distributions, respectively.

In a short period of time, this line of work has sparked great interest, and has led to remarkable results in very diverse applications -- ranging from the design optimization of light emitting diodes \cite{gomez2016design}, to the design of new molecules \cite{gomez2018automatic}, to the calibration of cosmological surveys \cite{ravanbakhsh2017enabling}, to RNA sequencing \cite{lopez2017deep}, to analyzing cancer gene expressions \cite{way2017extracting} -- all involving the approximation of very high-dimensional probability densities. It has also led to many fundamental studies that aim to further elucidate the capabilities and limitations of such models \cite{bousquet2017optimal,pu2017symmetric,rosca2018distribution,zheng2018degeneration,kingma2016improved, rezende2015variational}, enhance the interpretability of their results \cite{higgins2016beta,zhao2017infovae,chen2018isolating}, as well as  establish formal connections with well studied topics in mathematics and statistics, including importance sampling \cite{burda2015importance,klys2018joint} and optimal transport \cite{genevay2017gan, villani2008optimal, el2012bayesian}. 

In the original work of Kingma and Welling \cite{kingma2013auto} the encoder and decoder distributions, $q_{\phi}(\bm{z}|\bm{x},\bm{y})$ and  $p_{\theta}(\bm{y}|\bm{x},\bm{z})$, respectively, were both assumed to be Gaussian with a mean and a diagonal covariance that were parametrized  using feed-forward neural networks. Although this facilitates a straightforward evaluation of the lower bound in equation \ref{eq:ELBO}, it can result in a poor approximation of the true posterior $p(\bm{z}|\bm{x}, \bm{y})$ when the latter is non-Gaussian and/or multi-modal, as well as a poor reconstruction of the observed data \cite{rezende2015variational}. To this end, several methods have been proposed to overcome these limitations, including more expressive likelihood models \cite{van2016conditional}, more flexible variational approximations \cite{rezende2015variational,kingma2016improved,burda2015importance}, as well as reformulations that aim to make the variational bound of equation \ref{eq:ELBO} more tight \cite{liu2016stein,mescheder2017adversarial,makhzani2015adversarial,tolstikhin2017wasserstein,pu2017symmetric}. Overall, we must underline that such variational inference techniques are trading the rigorous asymptotic convergence guarantees that sampling-based methods like Markov Chain Monte Carlo enjoy, in favor of enhanced computational efficiency and performance, although new unifying ideas are aiming to bridge the gap between the two formulations \cite{titsias2017learning,blei2017variational}.
This trade-off becomes critical in tackling realistic large-scale problems, but it mandates careful validation of these tools to systematically assess  their performance.

In the next section we will revisit recent ideas in adversarial learning that enable us to overcome the limitations of classical mean field approximations \cite{wainwright2008graphical, blei2017variational}, and allow us to perform variational inference with arbitrarily flexible approximating distributions. These developments are unifying two of the most pioneering contributions in modern machine learning, namely variational auto-encoders and generative adversarial networks \cite{mescheder2017adversarial, pu2017symmetric,goodfellow2014generative}. Then, we will show how these techniques can be adapted to form the foundations of the proposed work, namely probabilistic data fusion and multi-fidelity modeling, and demonstrate how these tools can be used to accelerate the computational modeling of complex systems. 

\subsection{Adversarial learning with implicit distributions}
\label{sec:VAE}

The recent works of Pu {\em et. al.} \cite{pu2017symmetric} and Rosca {\em et. al.} \cite{rosca2018distribution} revealed some drawbacks in the original formulation of Kingma and Welling \cite{kingma2013auto} are attributed to the form of the variational objective in equation \ref{eq:ELBO}. Specifically, they showed that $\mathbb{KL}\left[q_{\phi}(\bm{z}|\bm{x})||p(\bm{z})\right]$ minimizes an upper bound on $\mathbb{KL}\left[q_{\phi}(\bm{z})||p(\bm{z})\right]$, where $q_{\phi}(\bm{z})=\int q_{\phi}(\bm{z}|\bm{x}) q(\bm{x})d\bm{x}$ is the marginal posterior over the latent variables $\bm{z}$, and $q(\bm{x})$ is the distribution of the observed data. 
By bringing $q_{\phi}(\bm{z})$ closer to
$p(\bm{z})$, the model distribution $p_{\theta}(\bm{x}) = \int
p_{\theta}(\bm{x}|\bm{z})p(\bm{z})d\bm{z}$ is
brought closer to the marginal reconstruction distribution
$\int p_{\theta}(\bm{x}|\bm{z})q_{\phi}(\bm{z})d\bm{z}$. Variational inference models learn to sample by maximizing reconstruction quality -- via the likelihood term $\mathbb{E}_{\bm{z}\sim q_{\phi}(\bm{z}|\bm{x})} \left[\log p_{\theta}(\bm{x}|\bm{z})\right]$ -- and reducing the gap between samples and
reconstructions -- via the $\mathbb{KL}$ term in equation \ref{eq:ELBO}. Failure to match $q_{\phi}(\bm{z})$
and $p(\bm{z})$ results in regions in latent space that have high
mass under $p(\bm{z})$ but not under $q_{\phi}(\bm{z})$. This means that prior samples $\bm{z} \sim p(\bm{z})$ passed through the decoder to obtain a model sample, are likely to be far in latent space from
inputs the decoder saw during training. It is this distribution
mismatch that results in poor generalization performance
from the decoder, and hence bad model samples.

\begin{figure}
\centering
\includegraphics[width=\textwidth]{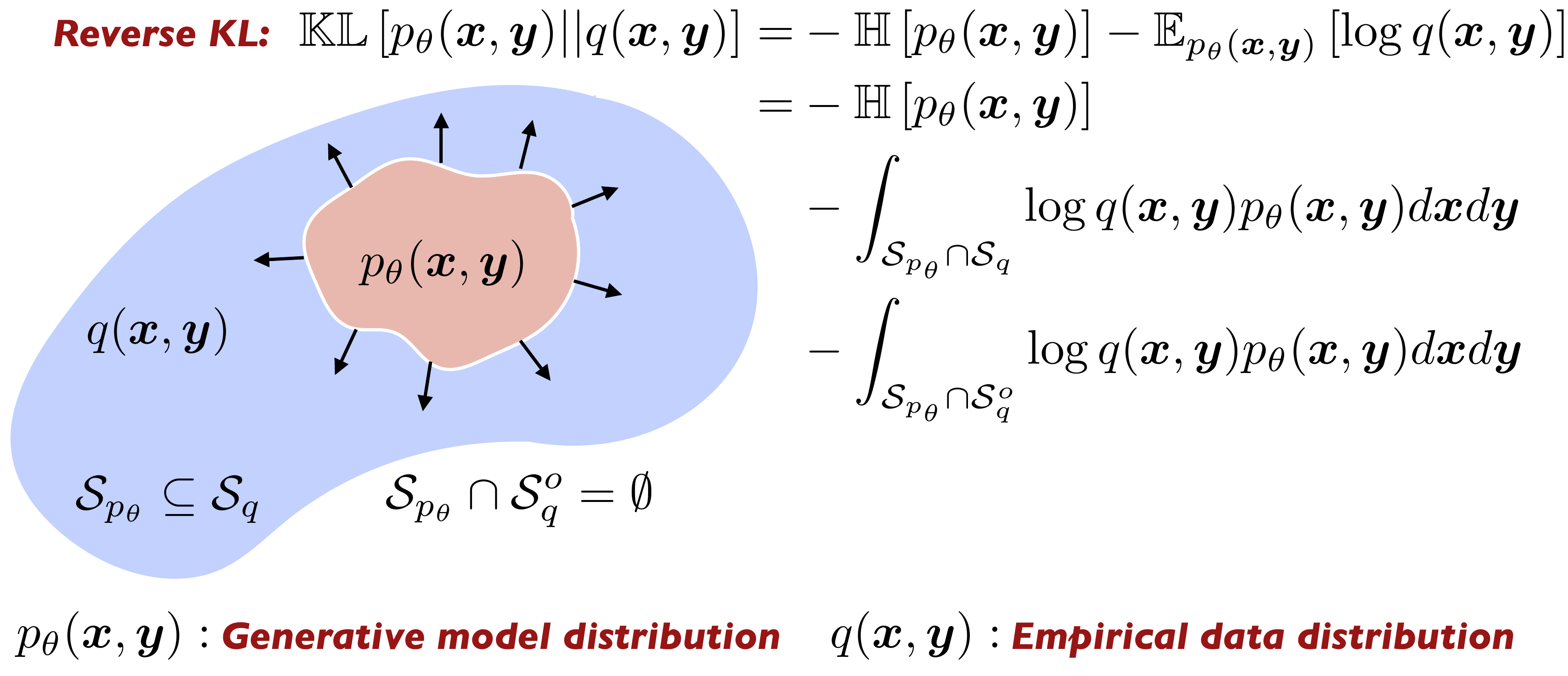}
\caption{{\em Joint distribution matching:} Schematic illustration of the proposed inference objective for joint distribution matching via minimization of the reverse KL-divergence. Penalizing a lower bound of the generative model entropy $\mathbb{H}(p_{\theta}(\bm{x},\bm{y})))$ provides a mechanism for mitigating the pathology of mode collapse in training adversarial generative models.}
\label{fig:joint_distribution_matching}
\end{figure}

Additional findings \cite{li2018learning} suggest that these  shortcomings can be overcome by introducing a new variational objective that aims to match the joint distribution of the generative model $p_{\theta}(\bm{x},\bm{y})$ with the joint empirical  distribution of the observed data $q(\bm{x},\bm{y})$. Matching the joint implies that that the respective marginal and conditional distributions are also encouraged to match. Here, we argue that matching the joint distribution of the generated data $p_{\theta}(\bm{x},\bm{y})$ with the joint distribution of the observed data $q(\bm{x},\bm{y})$ by  minimizing the reverse Kullback-Leibler divergence $\mathbb{KL}[p_{\theta}(\bm{x},\bm{y})||q(\bm{x},\bm{y})]$ is a promising approach to train the conditional generative model presented in equation \ref{eq:PSC}. To this end, the reverse Kullback-Leibler divergence reads as

\begin{align}\label{eq:KL}
\mathbb{KL}[p_{\theta}(\bm{x},\bm{y})||q(\bm{x},\bm{y})] & = -\mathbb{H}(p_{\theta}(\bm{x},\bm{y}))) - \mathbb{E}_{p_{\theta}(\bm{x},\bm{y})}[\log(q(\bm{x},\bm{y}))],
\end{align}
where $\mathbb{H}(p_{\theta}(\bm{x},\bm{y}))$ denotes the entropy of the conditional generative model. The second term can be further decomposed as

\begin{align}
\mathbb{E}_{p_{\theta}(\bm{x},\bm{y})}[\log(q(\bm{x},\bm{y}))]  = & \int_{\mathcal{S}_{p_{\theta}\cap \mathcal{S}_q}} \log(q(\bm{x},\bm{y})) p_{\theta}(\bm{x},\bm{y}) d\bm{x}d\bm{y} \ + \label{eq:decomposition} \\
& \int_{\mathcal{S}_{p_{\theta}\cap \mathcal{S}_q^{o}}} \log(q(\bm{x},\bm{y})) p_{\theta}(\bm{x},\bm{y}) d\bm{x} d\bm{y} \nonumber,
\end{align}
where $\mathcal{S}_{p_{\theta}}$ and $\mathcal{S}_q$ denote the support of the distributions $p_{\theta}(\bm{x},\bm{y})$ and $q(\bm{x},\bm{y})$, respectively, while $\mathcal{S}_q^{o}$ denotes the complement of $\mathcal{S}_q$. Notice that by minimizing the Kullback-Leibler divergence in equation \ref{eq:KL} we introduce a mechanism that is trying to balance the effect of two competing objectives. Specifically, maximization of the entropy term $\mathbb{H}(p_{\theta}(\bm{x},\bm{y})))$ encourages $p_{\theta}(\bm{x},\bm{y})$ to spread over its support set as wide, while the second integral term in equation \ref{eq:decomposition} introduces a strong (negative) penalty when the support of $p_{\theta}(\bm{x},\bm{y})$ and $q(\bm{x},\bm{y})$ do not overlap. Hence, the support of $p_{\theta}(\bm{x},\bm{y})$ is encouraged to spread only up to the point that $\mathcal{S}_{p_{\theta}}\cap \mathcal{S}_{q^{o}} = \emptyset$, implying that $\mathcal{S}_{p_{\theta}}\subseteq \mathcal{S}_{q^{o}}$. When $\mathcal{S}_{p_{\theta}}\subset \mathcal{S}_{q^{o}}$ the pathological issue of ``mode-collapse" (commonly encountered in the training of generative adversarial networks \cite{goodfellow2014generative}) is manifested \cite{salimans2016improved}. A visual sketch of this argument is illustrated in figure \ref{fig:joint_distribution_matching}.

The issue of mode collapse will also be present if one seeks to directly minimize the reverse Kullback-Leibler objective in equation \ref{eq:KL}, as this provides no control on the relative importance of the two
terms in the right hand side of equation \ref{eq:KL}. As discussed in \cite{li2018learning},  we may rather minimize $-\lambda \mathbb{H}(p_{\theta}(\bm{x},\bm{y}))) - \mathbb{E}_{p_{\theta}(\bm{x},\bm{y})}[\log(q(\bm{x},\bm{y}))]$, with $\lambda \ge 1$ to allow for control of how much
emphasis is placed on mitigating mode collapse. It is then clear that the entropic regularization introduced by $\mathbb{H}(p_{\theta}(\bm{x},\bm{y})))$ provides an effective mechanism for controlling and mitigating the effect of mode collapse, and, therefore, potentially enhancing the robustness adversarial inference procedures for learning $p_{\theta}(\bm{x},\bm{y})$.

Minimization of equation \ref{eq:KL} with respect to the generative model parameters $\theta$ presents two fundamental difficulties. First, the evaluation of both distributions $p_{\theta}(\bm{x},\bm{y})$ and $q(\bm{x},\bm{y})$ typically involves intractable integrals in high dimensions, and we may only have samples drawn from the two distributions, not their explicit analytical forms. Second, the differential entropy term $\mathbb{H}(p_{\theta}(\bm{x},\bm{y})))$ is intractable as $p_{\theta}(\bm{x},\bm{y}))$ is not known a-priori. In the next sections we revisit the unsupervised formulation put forth in \cite{li2018learning} and derive a tractable inference procedure for learning $p_{\theta}(\bm{x},\bm{y}))$ from scattered observation pairs $\{\bm{x}_{i}, \bm{y}_{i}\}$, $i = 1,\dots,N$.

\subsubsection{Density ratio estimation by probabilistic classification}\label{sec:density_ratio}

By definition, the computation of the reverse Kullback-Leibler divergence in equation \ref{eq:KL} involves computing an expectation over a log-density ratio, i.e. 
$$\mathbb{KL}[p_{\theta}(\bm{x},\bm{y})||q(\bm{x},\bm{y})] := \mathbb{E}_{p_{\theta}(\bm{x},\bm{y})}\left[\log\left(\frac{p_{\theta}(\bm{x},\bm{y})}{q(\bm{x},\bm{y})}\right)\right].$$ 
In general, given samples from two distributions, we can approximate their density ratio by constructing a binary classifier that distinguishes between samples from the two distributions. To this end, we assume that  $N$ data points are drawn from $p_{\theta}(\bm{x},\bm{y})$ and are assigned a label $c=+1$. Similarly, we assume that $N$ samples are drawn from  $q(\bm{x},\bm{y})$ and assigned label $c=-1$. Consequently, we can write these probabilities in a conditional form, namely
$$p_{\theta}(\bm{x},\bm{y}) = \rho(\bm{x},\bm{y}|c=+1), \ \ q(\bm{x},\bm{y}) = \rho(\bm{x},\bm{y}|c=-1),$$
where $\rho(\bm{x},\bm{y}|c=+1)$ and $\rho(\bm{x},\bm{y}|c=-1)$ are the class probabilities predicted by a binary classifier $T(\bm{x},\bm{y})$.
Using Bayes rule, it is then straightforward to show that the density ratio of $p_{\theta}(\bm{x},\bm{y})$ and $q(\bm{x},\bm{y})$ can be computed as

\begin{align}
    \frac{p_{\theta}(\bm{x},\bm{y})}{ q(\bm{x},\bm{y})} & = \frac{\rho(\bm{x},\bm{y}|c=+1)}{\rho(\bm{x},\bm{y}|c=-1)} \nonumber\\
    & = \frac{\rho(c=+1|\bm{x},\bm{y})\rho(\bm{x},\bm{y})}{\rho(c=+1)} \bigg/ \frac{\rho(c=-1|\bm{x},\bm{y})\rho(\bm{x},\bm{y})}{\rho(c=-1)} \nonumber\\
    & = \frac{\rho(c=+1|\bm{x},\bm{y})}{\rho(c=-1|\bm{x},\bm{y})} = \frac{\rho(c=+1|\bm{x},\bm{y})}{1 - \rho(c=+1|\bm{x},\bm{y})} \nonumber\\
    & = \frac{T(\bm{x},\bm{y})}{1-T(\bm{x},\bm{y})}.
\end{align}
This simple procedure suggests that we can harness the power of deep neural network classifiers to obtain accurate estimates of the reverse Kullback-Leibler divergence in equation \ref{eq:KL} directly from data and without the need to assume any specific parametrization for the generative model distribution $p_{\theta}(\bm{x},\bm{y})$.

\subsubsection{Entropic regularization bound}\label{sec:entropy_bound}
Here we follow the derivation of Li {\em et. al} \cite{li2018learning} to construct a computable lower bound for the entropy $\mathbb{H}(p_{\theta}(\bm{x},\bm{y}))$. To this end, we start by considering random variables $(\bm{x}, \bm{y}, \bm{z})$ under the joint distribution 
$$p_{\theta}(\bm{x}, \bm{y}, \bm{z}) = p_{\theta}(\bm{x}, \bm{y}|\bm{z}) p(\bm{z}) = p_{\theta}(\bm{y}|\bm{x}, \bm{z}) p(\bm{x}, \bm{z}),$$ 
where $p_{\theta}(\bm{y}|\bm{x}, \bm{z}) = \delta(\bm{y}-f_{\theta}(\bm{x}, \bm{z}))$, and $\delta(\cdot)$ is the Dirac delta function. The mutual information between $(\bm{x}, \bm{y})$ and $\bm{z}$ satisfies the information theoretic identity
$$ \mathbb{I}(\bm{x}, \bm{y}; \bm{z}) = \mathbb{H}(\bm{x}, \bm{y})-\mathbb{H}(\bm{x}, \bm{y}|\bm{z}) =
\mathbb{H}(\bm{z}) - \mathbb{H}(\bm{z}|\bm{x}, \bm{y}),$$
where $\mathbb{H}(\bm{x}, \bm{y})$, $\mathbb{H}(\bm{z})$ are the marginal entropies and $\mathbb{H}(\bm{x}, \bm{y}|\bm{z})$, $\mathbb{H}(\bm{z}|\bm{x}, \bm{y})$ are the conditional entropies \cite{akaike1998information}. 
Since in our setup $\bm{x}$ is a deterministic variables independent of $\bm{z}$, and samples of $p_{\theta}(\bm{y}|\bm{x}, \bm{z})$ are generated by a deterministic function $f_{\theta}(\bm{x}, \bm{z})$, it follows that $\mathbb{H}(\bm{x}, \bm{y}|\bm{z}) = 0$. 
We therefore have
\begin{equation}\label{eq:info_identity}
\mathbb{H}(\bm{x}, \bm{y}) = \mathbb{H}(\bm{z}) - \mathbb{H}(\bm{z}|\bm{x}, \bm{y}),
\end{equation}
where $\mathbb{H}(\bm{z}) := -\int \log p(\bm{z}) p(\bm{z}) d\bm{z}$ does not depend on the generative model parameters $\theta$. 

Now consider a general variational distribution $q_{\phi}(\bm{z}|\bm{x}, \bm{y})$ parametrized by a set of parameters $\phi$. Then,

\begin{align}
   \mathbb{H}(\bm{z}|\bm{x}, \bm{y})  = & -\mathbb{E}_{p_{\theta}(\bm{x}, \bm{y}, \bm{z})}[\log(p_{\theta}(\bm{z}|\bm{x}, \bm{y}))] \nonumber\\
    = & -\mathbb{E}_{p_{\theta}(\bm{x}, \bm{y}, \bm{z})}[\log(q_{\phi}(\bm{z}|\bm{x}, \bm{y}))] \nonumber  \\ 
    & -\mathbb{E}_{p_{\theta}(\bm{x}, \bm{y})}[\mathbb{KL}[p_{\theta}(\bm{z}|\bm{x}, \bm{y})||q_{\phi}(\bm{z}|\bm{x}, \bm{y})]] \nonumber \\
    \le & -\mathbb{E}_{p_{\theta}(\bm{x}, \bm{y}, \bm{z})}[\log(q_{\phi}(\bm{z}|\bm{x}, \bm{y}))]. \label{eq:entropy_bound}
\end{align}
Viewing $\bm{z}$ as a set of latent variables, then $q_{\phi}(\bm{z}|\bm{x}, \bm{y})$ is a variational approximation to the true intractable posterior over the latent variables $p_{\theta}(\bm{z}|\bm{x}, \bm{y})$. Therefore, 
if $q_{\phi}(\bm{z}|\bm{x}, \bm{y})$  is introduced as an auxiliary inference model associated with the generative model $p_{\theta}(\bm{x}, \bm{y})$, for which $\bm{y} = f_{\theta}(\bm{x}, \bm{z})$ and $\bm{z}\sim p(\bm{z})$, then we can use equations \ref{eq:info_identity} and \ref{eq:entropy_bound} to bound the entropy term in equation \ref{eq:KL} as
\begin{equation}
    \mathbb{H}(p_{\theta}(\bm{x},\bm{y})) \ge \mathbb{H}(p(\bm{z})) + \mathbb{E}_{p_{\theta}(\bm{x}, \bm{y}, \bm{z})}[\log(q_{\phi}(\bm{z}|\bm{x}, \bm{y}))].
\end{equation}
Note that the inference model $q_{\phi}(\bm{z}|\bm{x}, \bm{y})$ plays the role of a variational approximation to the true posterior over the latent variables, and appears naturally using information theoretic arguments in the derivation of the lower bound.

\subsubsection{Adversarial training objective}\label{sec:ADVI}

By leveraging the density ratio estimation procedure described in section \ref{sec:density_ratio} and the entropy bound derived in section \ref{sec:entropy_bound}, we can derive the following loss functions for minimizing the reverse Kullback-Leibler divergence with entropy regularization

\begin{align}
	\mathcal{L}_{\mathcal{D}}(\psi) = & \  \mathbb{E}_{q(\bm{x})p(\bm{z})}[\log\sigma(T_{\psi}(\bm{x},f_{\theta}(\bm{x},\bm{z})))] + \nonumber \\ & \ \mathbb{E}_{q(\bm{x},\bm{y})}[\log(1-\sigma(T_{\psi}(\bm{x},\bm{y})))] \label{eq:discriminator_loss}\\
	\mathcal{L}_{\mathcal{G}}(\theta, \phi) = & \ \mathbb{E}_{q(\bm{x},\bm{y})p(\bm{z})}[T_{\psi}(\bm{x}, f_{\theta}(\bm{x},\bm{z}))+ (1-\lambda)\log(q_{\phi}(\bm{z}|\bm{x},f_{\theta}(\bm{x},\bm{z}))) + \\ & \beta\|f_{\theta}(\bm{x},\bm{z}) - \bm{y}\|^2] \label{eq:generator_loss},
\end{align}
where $\sigma(x)=1/(1+e^{-x})$ is the logistic sigmoid function. For supervised learning tasks we can consider an additional penalty term controlled by the parameter $\beta$ that encourages a closer fit to the observed individual data points. Notice how the binary cross-entropy objective of equation \ref{eq:discriminator_loss} aims to progressively improve the ability of the classifier $T_{\psi}(\bm{x},\bm{y})$ to discriminate between ``fake" samples $(\bm{x},f_{\theta}(\bm{x},\bm{z}))$ obtained from the generative model $p_{\theta}(\bm{x},\bm{y})$ and ``true" samples $(\bm{x},\bm{y})$ originating from the observed data distribution $q(\bm{x},\bm{y})$. Simultaneously, the objective of equation \ref{eq:generator_loss} aims at improving the ability of the generator $f_{\theta}(\bm{x},\bm{y})$ to generate increasingly more realistic samples that can ``fool" the discriminator $T_{\psi}(\bm{x},\bm{y})$. Moreover, the encoder $q_{\phi}(\bm{z}|\bm{x},f_{\theta}(\bm{x},\bm{z}))$ not only serves as an entropic regularization term than allows us to stabilize model training and mitigate the pathology of mode collapse, but also provides a variational approximation to true posterior over the latent variables. The way it naturally appears in the objective of equation \ref{eq:generator_loss} also encourages the cycle-consistency of the latent variables $\bm{z}$; a process that is known to result in disentangled and interpretable low-dimensional representations of the observed data \cite{friedman2001elements}.

In theory, the optimal set of parameters $\{\theta^{\ast}, \phi^{\ast}, \psi^{\ast}\}$ correspond to the Nash equilibrium of the two player game defined by the loss functions in equations \ref{eq:discriminator_loss},\ref{eq:generator_loss}, for which one can show that the exact model distribution and the exact posterior over the latent variables can be recovered \cite{goodfellow2014generative, pu2017symmetric}. In practice, although there is no guarantee that this optimal solution can be attained, the generative model can be trained by alternating between optimizing the two objectives in equations \ref{eq:discriminator_loss},\ref{eq:generator_loss} using stochastic gradient descent as

\begin{align}
    & \mathop{\max}_{\psi} \ \mathcal{L}_{\mathcal{D}}(\psi) \label{eq:discriminator}\\
	& \mathop{\min}_{\theta, \phi} \ \mathcal{L}_{\mathcal{G}}(\theta, \phi) \label{eq:generator}.
\end{align}

\subsubsection{Predictive distribution}\label{sec:predictions}

Once the model is trained we can characterize the statistics of the outputs $\bm{y}$ by sampling latent variables from the prior $p(\bm{z})$ and passing them through the generator to yield conditional samples $\bm{y} = f_{\theta}(\bm{x}, \bm{z})$ that are distributed according to the predictive model distribution $p_{\theta}(\bm{y}|\bm{x})$. Note that although the explicit form of this  distribution is not known, we can efficiently compute any of its moments via Monte Carlo sampling. The cost of this prediction step is negligible compared to the cost of training the model, as it only involves a single forward pass through the generator function $f_{\theta}(\bm{x},\bm{z})$. Typically, we compute the mean and variance of the predictive distribution at a new test point $\bm{x}^{\ast}$ as 

\begin{align}
    \mu_{\bm{y}}(\bm{x}^{\ast}) & = \mathbb{E}_{p_{\theta}}[\bm{y}|\bm{x}^{\ast}, \bm{z}] \approx \frac{1}{N_s}\sum\limits_{i=1}^{N_s} f_{\theta}(\bm{x}^{\ast},  \bm{z}_i), \label{eq:predictive_mean} \\
    \sigma^{2}_{\bm{y}}(\bm{x}^{\ast}) & = \mathbb{V}\text{ar}_{p_{\theta}}[\bm{y}|\bm{x}^{\ast}, \bm{z}] \approx \frac{1}{N_s}\sum\limits_{i=1}^{N_s} [f_{\theta}(\bm{x}^{\ast}, \bm{z}_i) - \mu_{\bm{y}}(\bm{x}^{\ast})]^2, \label{eq:predictive_variance}
\end{align}
where $\bm{z}_i \sim p(\bm{z})$, $i = 1,\dots,N_s$, and $N_s$ corresponds to the total number of Monte Carlo samples.

\section{Results}\label{sec:results}

Here we present a diverse collection of demonstrations to showcase the broad applicability of the proposed methods.
Moreover, in  \ref{sec:appendix} we provide a comprehensive collection of systematic studies that aim to elucidate the robustness of the proposed algorithms with respect to different parameter settings.  
In all examples presented in this section we have trained the models for 20,000 stochastic gradient descent steps using the Adam optimizer \cite{kingma2014adam} with a learning rate of $10^{-4}$, while fixing a one-to-five ratio for the discriminator versus generator updates. Unless stated otherwise, we have also fixed the entropic regularization and the residual penalty parameters to $\lambda = 1.5$ and $\beta = 0.0$, respectively. The proposed algorithms were implemented in Tensorflow v1.10 \cite{abadi2016tensorflow}, and computations were performed in single precision arithmetic on a single NVIDIA Tesla P100 GPU card. All data and code accompanying this manuscript will be made available at \url{https://github.com/PredictiveIntelligenceLab/CADGMs}.

\subsection{Regression of noisy data}\label{sec:regression}
We begin our presentation with an example in which the observed data is generated by a deterministic process but the observations are stochasticaly perturbed by random noise. Specifically, we consider the following three distinct cases: 
\begin{enumerate}[label=(\roman*)]
    \item {\em Gaussian homoscedastic noise:}
    \begin{equation}
        g(x) = \log(10 (|x-0.03|+0.03))\sin(\pi (|x-0.03|+0.03)) + \delta,
    \end{equation} 
    where $\delta$ corresponds to $5\%$ uncorrelated zero-mean Gaussian noise.
    \item {\em Gaussian heteroscedastic noise:}
    \begin{equation}
        g(x) = \log(10 (|x-0.03|+0.03))\sin(\pi (|x-0.03|+0.03)) + \delta(x), 
    \end{equation} 
    where $\delta(x) = \frac{\epsilon}{\exp (2(|x-0.03|+0.03))}$, and  $\epsilon\sim N(0, 0.5^2)$. 
    \item {\em Non-additive, non-Gaussian noise:}
    \begin{equation}
    g(x) = \log(10 (|x-0.03|+0.03))\sin(\pi (|x-0.03|+0.03) + 2\delta(x)) + \delta(x)
    \end{equation} 
    where $\delta(x) = \frac{\epsilon}{\exp (2(|x-0.03|+0.03))}$, and $\epsilon\sim N(0, 0.5^2)$.
\end{enumerate}
In all cases, we assume access to $N=200$ training pairs $\{\bm{x}_i, \bm{y}_i\}$, $i=1,\dots,N$ randomly sampled in the interval $x\in[-2,2]$ according to the empirical data distribution $q(\bm{x},\bm{y})$. Then, our goal is to approximate the conditional distribution $p_{\theta}(\bm{y}|\bm{x},\bm{z})$ using a generative model $\bm{y} = f_{\theta}(\bm{x}, \bm{z})$, $\bm{z}\sim p(\bm{z})$, that combines the original inputs $\bm{x}$ and a set of latent variables $\bm{z}$ to predict the outputs $\bm{y}$.

As described in section \ref{sec:Methods}, the outputs $\bm{y}$ are generated by pushing the inputs $\bm{x}$ and the latent variables $\bm{z}$ through a deterministic  generator function $f_{\theta}(\bm{x},\bm{z})$, typically parametrized by deep neural networks. Moreover, a discriminator network  is used to minimize the reverse KL-divergence between the generative model distribution $p_{\theta}(\bm{x},\bm{y})$ and the empirical data distribution $q(\bm{x},\bm{y})$. Finally, we introduce an auxiliary inference network to model the approximate posterior distribution over the latent variables, namely $q_{\phi}(\bm{z}|\bm{x}, \bm{y})$ that encodes the observed data $(\bm{x}, \bm{y})$ into a latent space using a deterministic mapping $\bm{z} = f_{\phi}(\bm{x}, \bm{y})$, also modeled using a deep neural network.

The proposed conditional generative model is constructed using fully connected feed-forward architectures for the encoder and generator networks with 3 hidden layers and 100 neurons per layer, while the discriminator architecture has 2 hidden layers with 100 neurons per layer. All activation use a hyperbolic tangent non-linearity, and we have not employed any additional modifications such as L2 regularization, dropout or batch-normalization \cite{goodfellow2016deep}. During model training, for each epoch we train the discriminator for two times, and encoder and generator for one time using stochastic gradient updates with the Adam optimizer \cite{kingma2014adam} and a learning rate of $10^{-4}$ using the entire data batch. Finally, we set the entropic regularization penalty parameter $\lambda = 1.5$.

\begin{figure}
\centering
\includegraphics[width=\textwidth]{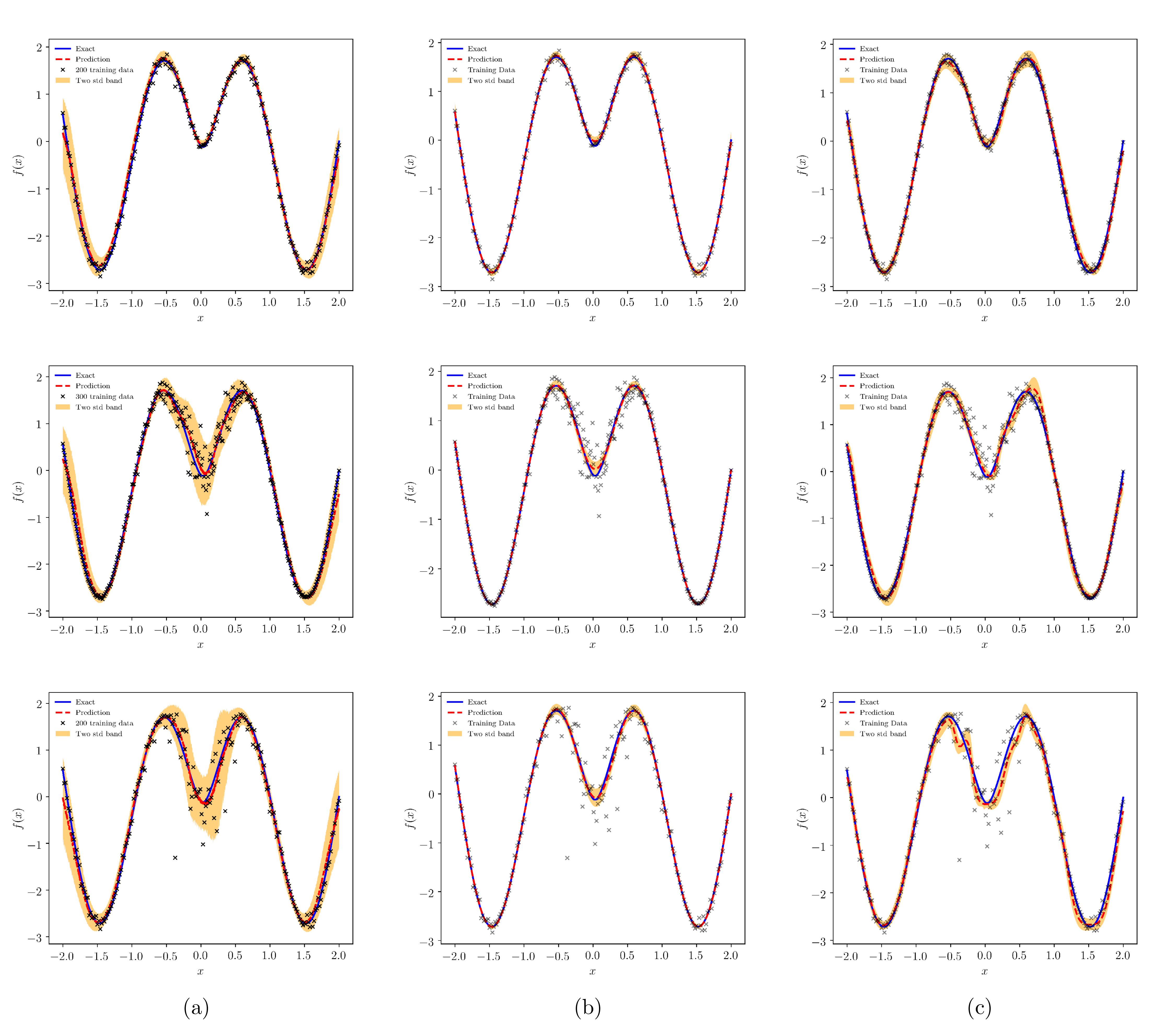}
\caption{{\em Regression under homoscedastic noise:} Training data (black crosses) and the exact noise-free solution (blue solid line) versus the predictive mean (red dashed line)and two standard deviations (orange shaded region) obtained by: (a) the proposed conditional generative model, (b) a Gaussian process regression model, and (c) a Bayesian neural network. {\it Top row panels:} Gaussian homoscedastic noise, {\it Middle row panels:} Gaussian heteroscedastic noise, {\it Bottom row panels:} Non-additive, non-Gaussain noise.}
\label{fig:regression}
\end{figure}

Figure \ref{fig:regression} summarizes our results for all cases  obtained using:
\begin{enumerate}[label=(\alph*)]
    \item The proposed conditional generative model described above.
    \item A simple Gaussian process model with a Gaussian likelihood and a squared exponential covariance function trained using exact inference \cite{rasmussen2004gaussian}.
    \item A Bayesian neural network having the same architecture as the generator network described above and trained using mean-field stochastic variational inference \cite{neal2012bayesian}
\end{enumerate}
 We observe that the proposed  conditional generative model returns robust predictions with sensible uncertainty estimates for all cases. On the other hand, the basic Gaussian process and Bayesian neural network models perform equally well for the simple uncorrelated  noise case, but suffer from over-fitting and fail to return reasonable uncertainty estimates for the more complex heteroscedastic and non-additive cases. These predictions could in principle be improved with the use of more elaborate priors, likelihoods and inference procedures, however such remedies often hamper the practical applicability of these methods. In contrast, the proposed conditional generative model appears to be robust across these inherently different cases without requiring any modifications or specific assumptions regarding the nature of the noise process.

\subsection{Multi-fidelity modeling of stochastic processes}
\label{sec:multifidelity}
In this section we demonstrate how the proposed methodology can be adapted to accommodate the setting of supervised learning from data of variable fidelity. Let it be a synthesis of expensive experiments and simplified analytical models, multi-scale/multi-resolution computational models, or historical data and expert opinion, the concept of multi-fidelity modeling lends itself to enabling effective pathways for accelerating the analysis of systems that are prohibitively expensive to evaluate. As discussed in section \ref{sec:Introduction}, these methods have been successful in a wide spectrum of applications including design optimization \cite{forrester2007multi, robinson2008surrogate, alexandrov2001approximation, sun2010two, sun2011multi}, model calibration \cite{perdikaris2016model, perdikaris2015data, perdikaris2015calibration}, and uncertainty quantification \cite{eldred2009comparison, ng2012multifidelity,padron2014multi, biehler2015towards, peherstorfer2016optimal, peherstorfer2016multifidelity, peherstorfer2016survey, narayan2014stochastic, zhu2014computational, bilionis2013multi, parussini2017multi,perdikaris2016multifidelity}. 

Except perhaps for Gaussian process regression models, most existing approaches to multi-fidelity modeling are trying to construct deterministic surrogates of some form $\bm{y}=f(\bm{x})$, and use theoretical error bounds to quantify the accuracy of the surrogate model predictions. For instance, a multi-fidelity problem can be formulated by considering $\bm{y}:=\{y_l\}$ and $\bm{x}:=\{x,\lambda,y_1,y_2,\dots,y_{l-1},\}$, where $y_l$ is the output of our highest fidelity information source, $(y_1,y_2,\dots,y_{l-1})$ are predictions of lower fidelity models, $x$ is a vector of space-time  coordinates, and $\lambda$ is a vector of uncertain parameters. Despite their growing popularity, the applicability of multi-fidelity modeling techniques is typically limited to systems that are governed by deterministic input-output relations. To the best of our knowledge, this is the first attempt of applying the concept of multi-fidelity modeling to expedite the statistical characterization of correlated stochastic processes.

Without loss of generality, and to keep our presentation clear, we will focus on a setting involving two correlated stochastic processes. Intuitively, one can think of the following example scenario. We want to characterize the statistics of a random quantity of interest (e.g., velocity fluctuations of a turbulent flow near a wall boundary) by recording its value at a finite set of locations and for a finite number of random realizations. However, these recordings may be hard/expensive to obtain as they may require a set of sophisticated and well calibrated sensors, or a set of fully resolved computational simulations. At the same time, it might be easier to obtain more measurements either by probing the same quantity of interest using a set of cheaper/uncalibrated sensors (or simplified/coarser computational models), or by probing an auxiliary quantity of interest that is statistically correlated to our target variable but is much easier to record (e.g., sensor measurements of pressure on the wall boundary). Then our goal is to synthesize these measurements and construct a predictive model that can fully characterize the statistics of the target stochastic process.

More formally, we assume that we have access to a number of {\em high-fidelity} input-output pairs $(\bm{x}_H,\bm{y}_H)$ corresponding to a finite number of realizations of the target stochastic process, measured at a handful input locations $x_H$ using high-fidelity sensors. Moreover, we also have access to {\em low-fidelity}  input-output pairs $(\bm{x}_L,\bm{y}_L)$ corresponding to a finite number of realizations of either the target stochastic process or an auxiliary process that is statistically correlated with the target, albeit probed for a much larger collection of inputs. Then our goal is to learn the conditional distribution $p_{\theta}(\bm{y}_H|\bm{x}_H, \bm{y}_L, \bm{z})$ using a generative model $\bm{y}_H = f_{\theta}(\bm{x}_H, \bm{y}_L, \bm{z})$, $\bm{z}\sim p(\bm{z})$.

\begin{figure}
\centering
\includegraphics[width=\textwidth]{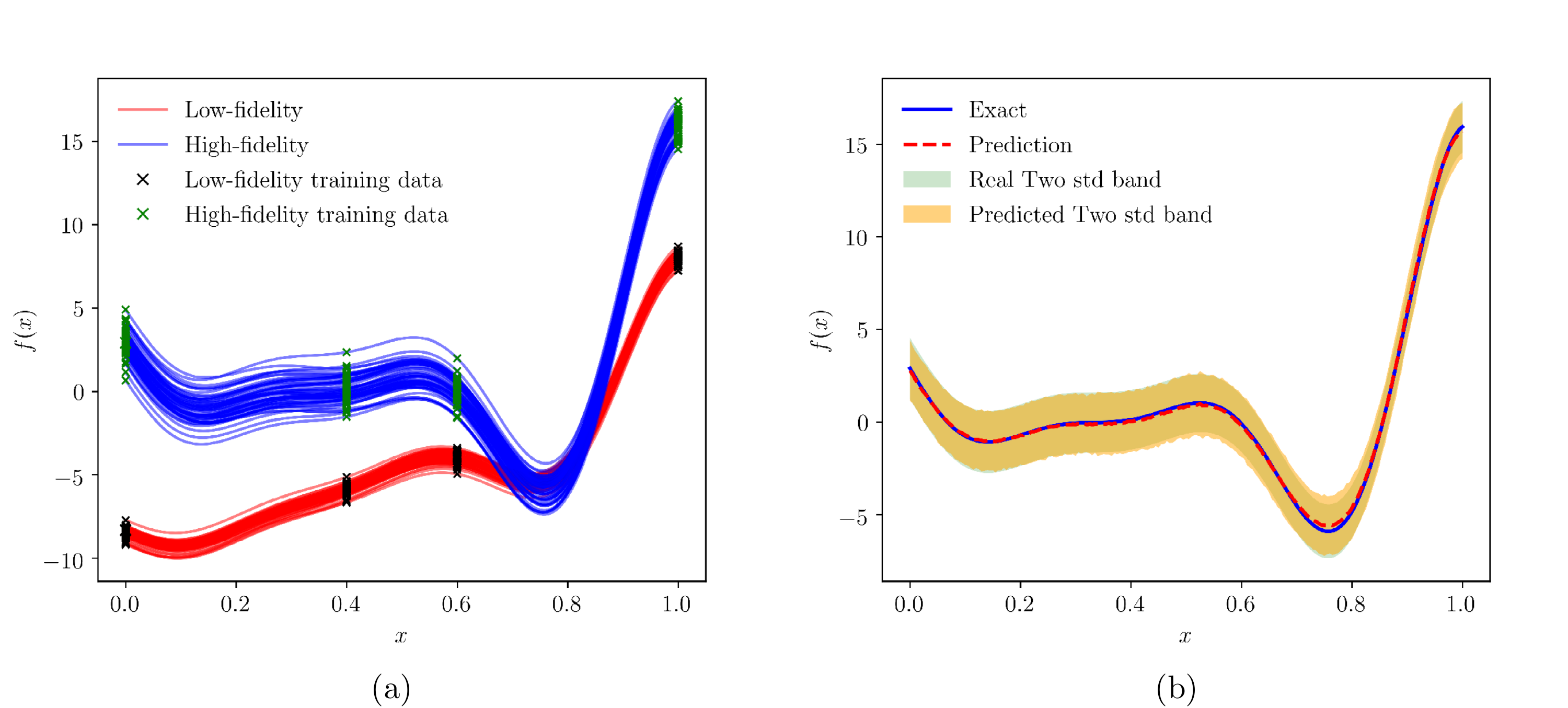}
\caption{{\em Multi-fidelity modeling of stochastic processes:} (a) Sample realizations of the low- and high-fidelity stochastic processes (red and blue lines, respectively) along with the sensor measurements at $x = [0, 0.4, 0.6, 1.0]$ used to train the generative model (black and green crosses, respectively). (b) Predicted mean (red dashed line) and two standard deviations (yellow band) for the high-fidelity stochastic process versus the exact solution (blue solid line and green band, respectively).}
\label{fig:MuFi}
\end{figure}

We will illustrate this work-flow using a synthetic example involving data generated from two correlated Gaussian processes in one input dimension

\begin{align}
\left[ \begin{array}{c} f_{L}(x) \\ f_{H}(x) \end{array} \right] 
& \sim \mathcal{N}\left(\left[\begin{array}{c} \mu_L(x) \\ \mu_H(x) \end{array} \right],  
\left[ \begin{array}{c c} K_{LL} & K_{LH}
 \\ K_{LH}' & K_{HH}
\end{array} \right]\right), \\
\end{align}
with a mean and covariance functions given by
\begin{align}
    \mu_L(x) &= 0.5\mu_H(x) + 10(x-0.5) - 5 \\ 
	\mu_H(x) &= (6x-2)^2\sin(12x-4) \label{eq:mu_H}\\
K_{LL} & =  k(x,x;\theta_{L})  \\
K_{LH} & = \rho k(x,x;\theta_{L}) \\
K_{HH} & = \rho^2 k(x,x;\theta_{L}) + k(x,x;\theta_{H}).
\end{align}
Here $\theta_{L}=(\sigma_{f_L}^2, l_L^2)$ and $\theta_{H}=(\sigma_{f_H}^2, l_H^2)$ correspond to two different sets of hyper-parameters of a square exponential kernel
\begin{equation}\label{eq:rbf_kernel}
k(x,x';\theta) = \sigma_f^2\exp\left(-\frac{(x-x')^2}{2l^2}\right).
\end{equation}
Moreover, $\rho$ is a parameter that controls the degree to which the two stochastic processes exhibit linear correlations \cite{kennedy2000predicting, perdikaris2017nonlinear}.  In this example we have considered $\sigma_{f_L}^2 = 0.1, l_L^2 = 0.5, \sigma_{f_H}^2 = 0.5, l_H^2=0.5$ and $\rho=0.8$, and generated a training data-set consisting of 50 realizations of $f_L(x)$ and $f_H(x)$ recorded using a set of sensors fixed at locations $x_L = x_H = [0, 0.4, 0.6, 1.0]$ (see figure \ref{fig:MuFi}(a)). 

We employ a conditional generative model constructed using simple feed-forward neural networks with 3 hidden layers and 100 neurons per layer for both the generator and the encoder, and 2 hidden layers with 100 neurons per layer for the discriminator. The activation function in all cases is chosen to be a hyperbolic tangent non-linearity. Moreover, we have chosen a one-dimensional latent space with a standard normal prior, i.e. $z\sim\mathcal{N}(0,1)$. Model training is performed using the Adam optimizer \cite{kingma2014adam} with a learning rate of $10^{-4}$ for all the networks. For each stochastic gradient descent iteration, we update the discriminator for 1 time and the generator for 5 times, while we fix the entropic regularization penalty parameter to $\lambda = 1.5$. Notice that during model training the algorithm only requires to see joint observations of $f_L(x)$ and $f_H(x)$ at a fixed set of input locations $x$ (see figure \ref{fig:MuFi}(a)). However, during prediction at a new test point $x^{\ast}$ one needs to first sample $y_L^{\ast} = f_L(x^{\ast})$, and then use the generative model to produce samples $y_H^{\ast} = f_{\theta}(x^{\ast}, y_L^{\ast}, z)$, $z\sim\mathcal{N}(0,1)$. 

\begin{figure}
\centering
\includegraphics[width=\textwidth]{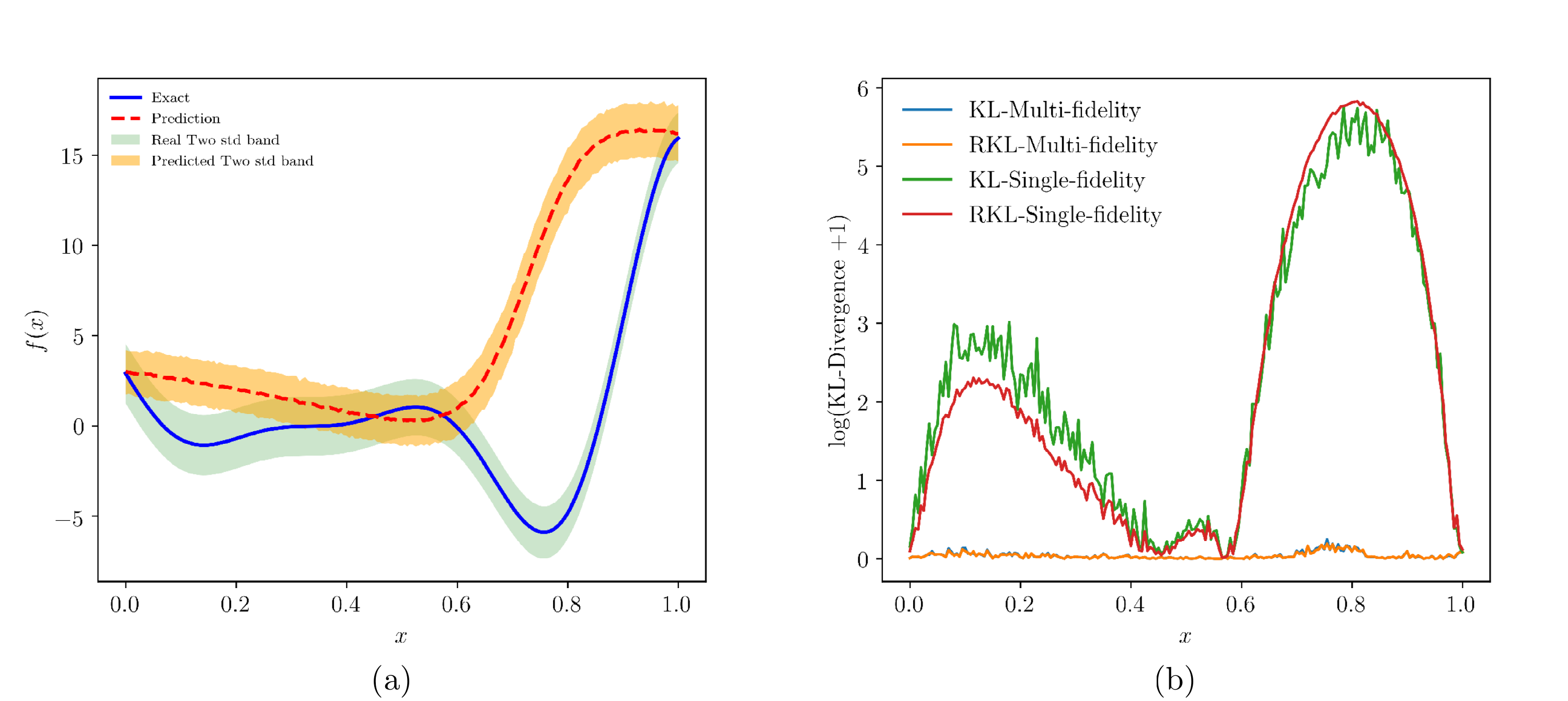}
\caption{{\em Multi-fidelity modeling of stochastic processes:} (a)
Predicted mean (red dashed line) and two standard deviations (yellow band) of a single-fidelity conditional generative model versus the exact solution (blue solid line and green band, respectively). (b)
Comparison of the KL-divergence and Reverse-KL-divergence between the exact marginal densities and the predictions of the single- and  multi-fidelity conditional generative models.}
\label{fig:MuFi_KL}
\end{figure}

The results of this experiment are summarized in figures \ref{fig:MuFi}(b) and \ref{fig:MuFi_KL}. Specifically, in figure \ref{fig:MuFi}(b) we observe a qualitative agreement between the second order sufficient statistics for the predicted and the exact  high-fidelity processes. The effectiveness of the multi-fidelity approach becomes evident when we compare our results against a single-fidelity conditional generative model trained only on the high-fidelity data. The result of this experiment is presented in \ref{fig:MuFi_KL}(a) where it is clear that the generative model fails to correctly capture the target stochastic process. To make this comparison quantitative, we have estimated the forward and reverse Kullback-Leibler divergence for a collection of one-dimensional marginal distributions corresponding to different spatial locations in $x\in[0,1]$. To this end, we have employed a Gaussian approximation for the predicted marginal densities of the generative model and compared them against the exact Gaussian marginal densities of the target high-fidelity process using the analytical expression for the KL-divergence between two Gaussian distributions
$p_1(x)\sim\mathcal{N}(\mu_1, \sigma_1^2)$ and
$p_2(x)\sim\mathcal{N}(\mu_2, \sigma_2^2)$,
\begin{align}
    \mathbb{KL}[p_1(x)||p_2(x)] &= - \int p_1(x)\log p_2(x) dx + \int p_1(x)\log p_1(x) dx \nonumber \\ 
	&= \frac{1}{2}\log(2\pi\sigma_2^2) + \frac{\sigma_1^2 + (\mu_1 - \mu_2)^2}{2\sigma_2^2} - \frac{1}{2}(1 + \log(2\pi\sigma_1^2)) \nonumber \\
    & = \log\frac{\sigma_2}{\sigma_1} + \frac{\sigma_1^2 + (\mu_1 - \mu_2)^2}{2\sigma_2^2} - \frac{1}{2}. \label{eq:KL_div}
\end{align}
The result of this comparison is shown in figure \ref{fig:MuFi_KL}(b) for both the single- and multi-fidelity cases. Clearly, the appropriate utilization of the low-fidelity data results in significant accuracy gains for the multi-fidelity case, while the single-fidelity model is not able to generalize well and suffers from large errors in KL-divergence in all locations away from the training data.

\subsection{Uncertainty propagation in high-dimensional dynamical systems}
\label{sec:highdimUQ}
\begin{figure}
\centering
\includegraphics[width=0.6\textwidth]{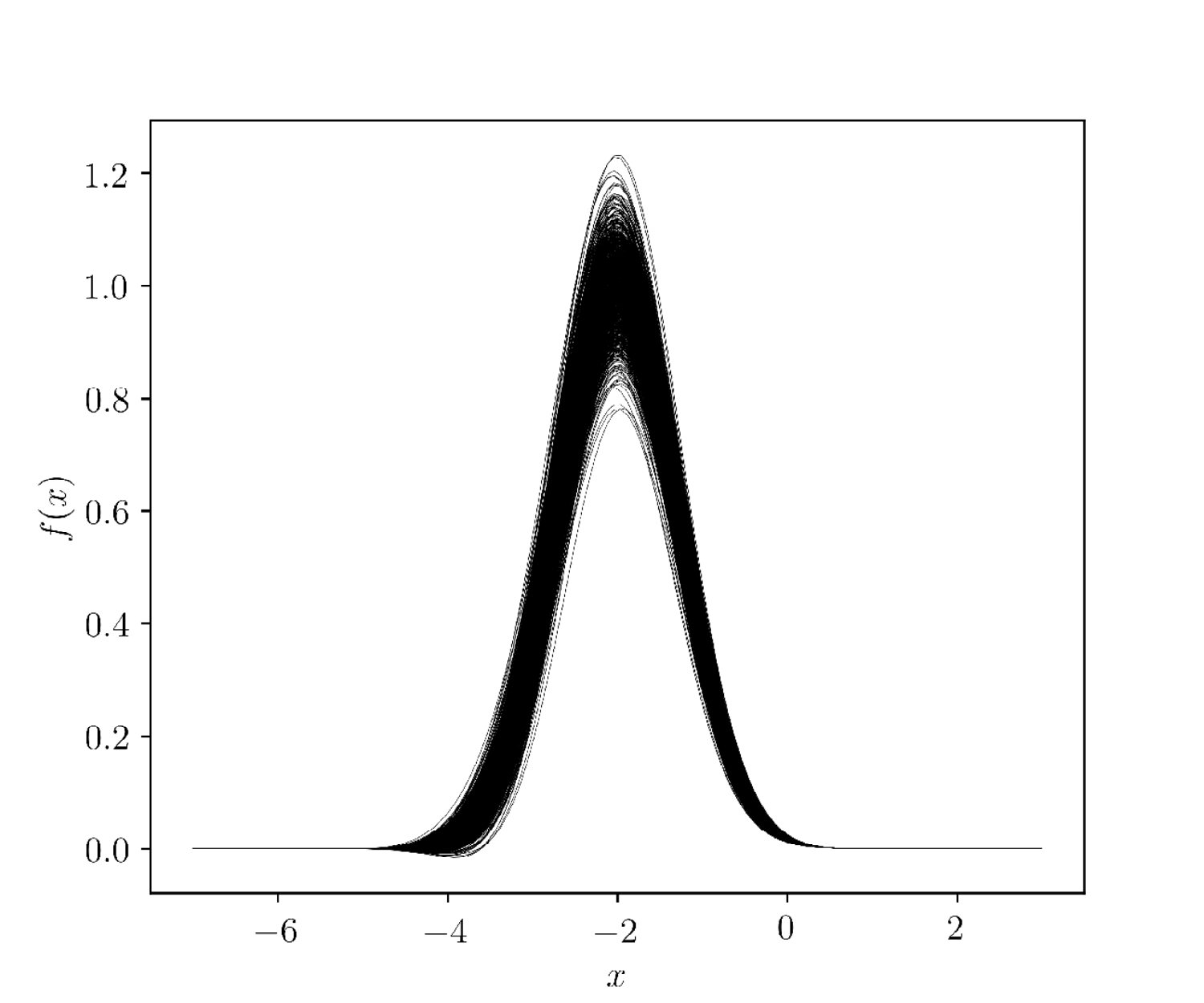}
\caption{{\em Uncertainty propagation in high-dimensional dynamical systems:} 100 representative samples of a conditional Gaussian process used as initial conditions for the Burgers equation.}
\label{fig:Burgers_init}
\end{figure}

In this section we aim to demonstrate how the proposed inference framework can leverage modern deep learning techniques to tackle high-dimensional uncertainty propagation problems involving complex dynamical systems. To this end, we will consider the temporal evolution of the non-linear time-dependent Burgers equation in one spatial dimension, subject to random initial conditions. The equation and boundary conditions read as
\begin{equation}
\label{eq:Burgers}
\begin{aligned}
	&u_t + u u_x - \nu u_{xx} = 0, \quad\quad x\in[-7, 3], t\in[0,50],\\
	&u(t,-7) = u(t,3) = 0,\\
\end{aligned}
\end{equation}
where the viscosity parameter is chosen as $\nu = 0.5$ \cite{burgers1948mathematical}. We will evolve the system starting from a random initial condition generated by a conditional Gaussian process \cite{rasmussen2004gaussian} that constrains the initial sample paths to satisfy zero Dirichlet boundary conditions, i.e. $u(0,x)\sim\mathcal{GP}(\mu(x), \Sigma(x))$, with
\begin{equation}
\label{eq:Burgers_initial}
\begin{aligned}
	\mu(x) &= k(x,x_b)K^{-1}y_b, \\
	\Sigma(x) &= k(x,x) - k(x,x_b)K^{-1}k(x_b,x),\\ 
\end{aligned}
\end{equation}
where $x_b$ and $y_b$ are column vectors corresponding to zero data near the domain boundaries, and $K$ is a covariance matrix constructed by evaluating the square exponential kernel (see equation \ref{eq:rbf_kernel}) with fixed variance and length-scale hyper-parameters $\sigma_f^2 = 0.005$ and $l^2 = 1$, respectively (see figure \ref{fig:Burgers_init}). The resulting solution to this problem is a continuous spatio-temporal random field $u(x,t)$ whose statistical description defines a non-trivial infinite-dimensional uncertainty propagation problem. As we will describe below, we will leverage the capabilities of convolutional neural networks in order to construct a scalable surrogate model that is capable of providing a complete statistical characterization of the random field $u(x,t)$ for any time $t$ and for a finite collection of spatial locations $x$.

 We generate a data-set consisting of 100 sample realizations of the system in the interval $t\in [0, 50]$ using a high-fidelity Fourier spectral method \cite{kassam2005fourth} on a regular spatial grid consisting of 128 points and 256 time-steps. Our goal here is to use a subset of this data to train a deep generative model for approximating the conditional density $p_{\theta}(\bm{u}|t,\bm{z})$, $\bm{z}\sim p(\bm{z})$, where the vector $\bm{u}\in\mathbb{R}^{128}$ corresponds to the collocation of the continuous field $u(x,t)$ at the 128 spatial grid-points for a given temporal snapshot at time $t$. We use data from 64 randomly selected temporal snapshots to train the generative model, and the rest will be used for validating our results. 



To exploit the gridded structure of the data we employ 1d-convolutional neural networks \cite{krizhevsky2012imagenet} which allow us to construct a multi-resolution representation of the data that can capture local spatial correlations \cite{lecun2015deep,mallat2016understanding}. To this end, the generator network is constructed using 5 transposed convolution layers with channel sizes of $[512, 256, 128, 64, 32]$, kernel size 4, stride 2, and a hyperbolic tangent activation function in all layers except the last. For the encoder we use 5 convolutional layers with channel sizes of $[32, 64, 128, 128, 256]$, each with a kernel size of 5, stride 2, followed by a batch normalization layer \cite{ioffe2015batch} and a hyperbolic tangent activation. The last layer of the encoder is a fully connected layer that returns outputs with the same dimension of $\bm{z}$. Here, we choose the latent space  dimension to be 32, i.e. $\bm{z}\in \mathbb{R}^{32}$, with an isotropic normal prior, $p(\bm{z})\sim\mathcal{N}(\bm{0},\bm{I})$. Finally, for the discriminator we use 4 convolution layers with the channel sizes of $[32, 64, 128, 256]$, each with kernel size of 5, stride 2, and a hyperbolic tangent activation function in all layers except the last. The last layer of the discriminator is a fully connected layer to convert the final output into scalar class probability predictions that aim to correctly distinguish between real and generated samples in the 128-dimensional output space.

Notice that the time variable is treated as a continuous label corresponding to each time instant $t$, and it is incorporated in our work-flow as follows. For the discriminator and the encoder, we broadcast time as a vector having the same size of the data and treat it as an additional input channel. For the decoder, we broadcast time as a vector having the same size of the latent variable and concatenate them together. We use the Adam \cite{kingma2014adam} optimizer with the learning rate $10^{-4}$ for all the networks. For each epoch, we train the discriminator for 1 time and the generator for 1 time. Finally, we set the entropic regularization penalty to $\lambda = 1.5$ and the data fit penalty to $\beta = 0.5$ (see equation \ref{eq:generator_loss}).

\begin{figure}
\centering
\includegraphics[width=\textwidth]{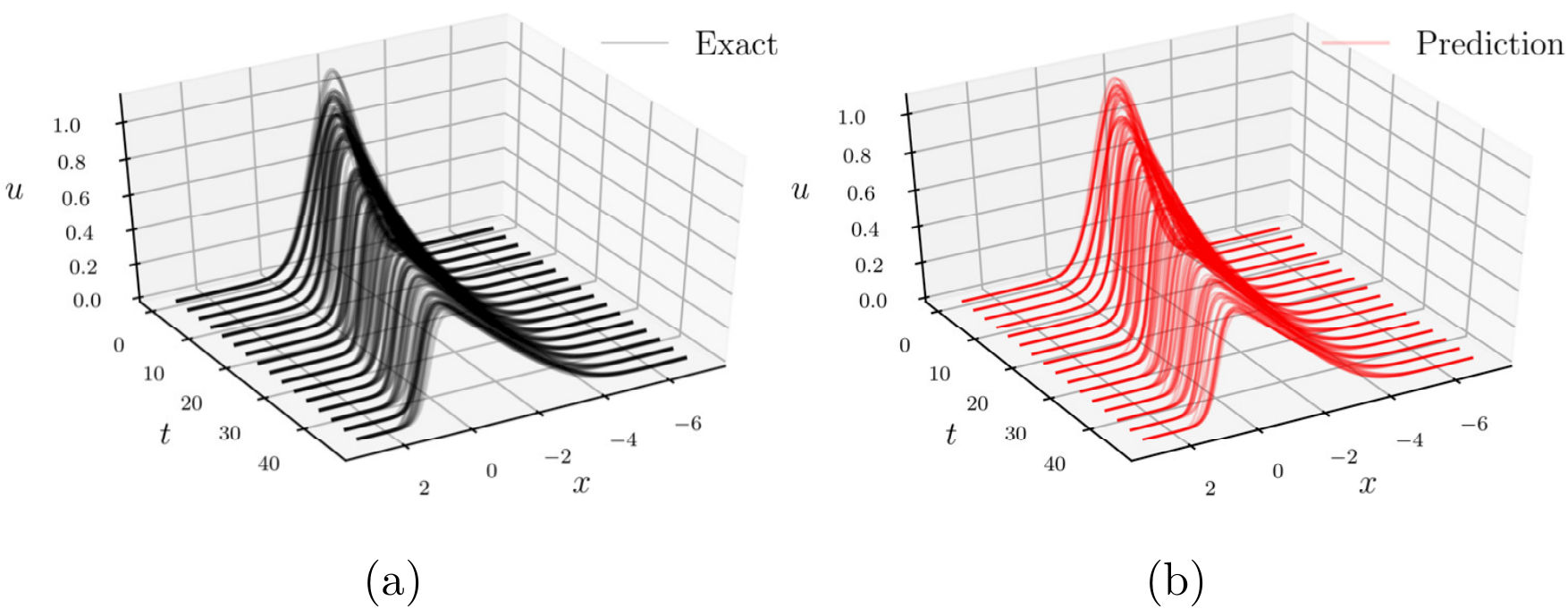}
\caption{{\em Uncertainty propagation in high-dimensional dynamical systems:} (a) Exact sample trajectories of the Burgers equation. (b) Samples generated by the conditional generative model $p_{\theta}(\bm{u}|t,\bm{z})$. The comparison corresponds to 16 different temporal snapshots and depicts 10 samples per snapshot. Each sample is a 128-dimensional vector.}
\label{fig:Burgers_samples}
\end{figure}

\begin{figure}
\centering
\includegraphics[width=\textwidth]{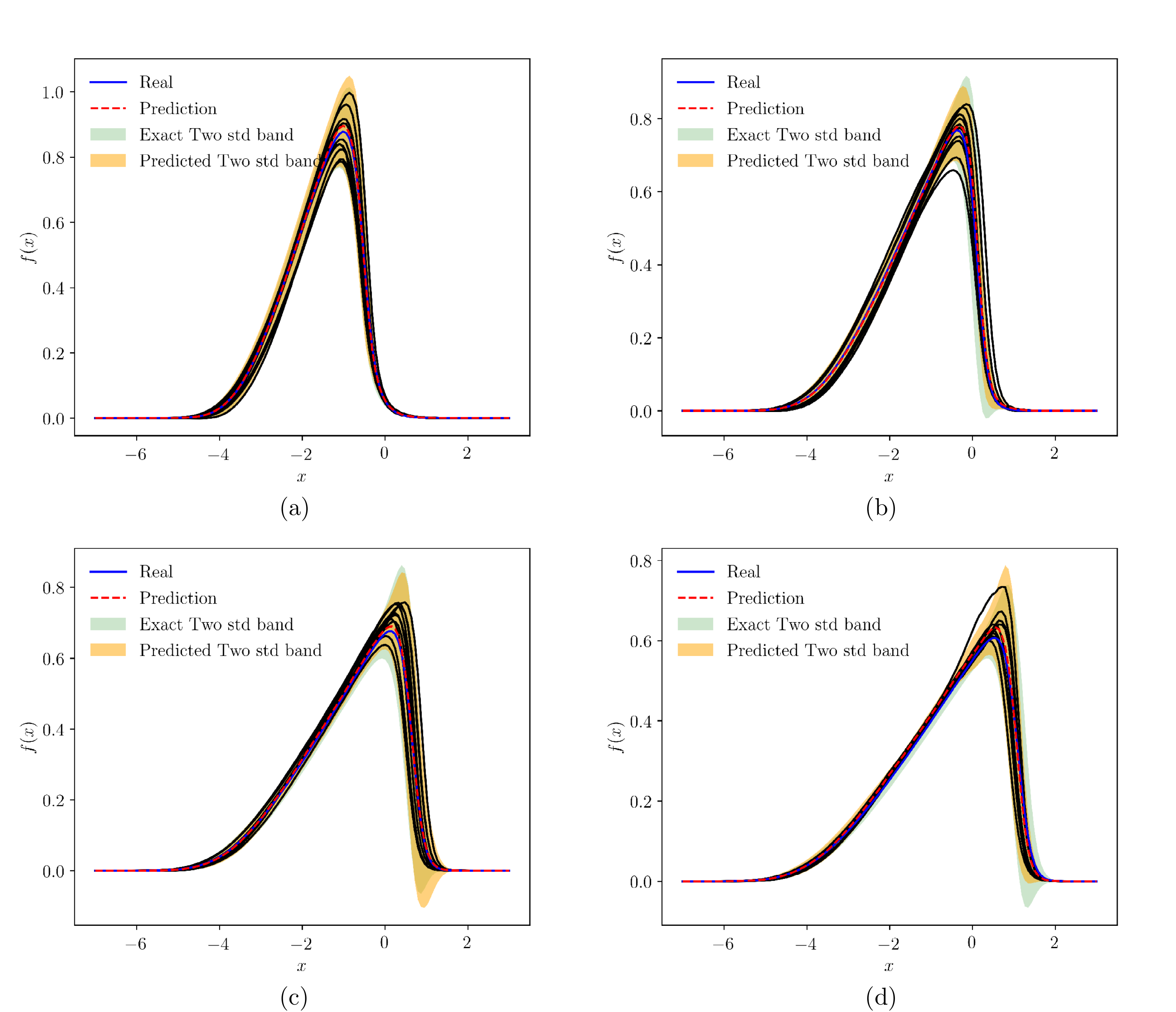}
\caption{{\em Uncertainty propagation in high-dimensional dynamical systems:} Mean (solid blue line) and two standard deviations (green shaded region) of reference simulated trajectories of the Burgers equation versus the predictions of the conditional generative model $p_{\theta}(\bm{u}|t,\bm{z})$ (red dashed line and yellow shaded region, respectively). Results are reported for four temporal instances  that were not used during model training: (a) $t = 12.5$, (b) $t = 25$, (c) $t = 37.5$, and (d) $t = 50$.}
\label{fig:Burgers_marginals}
\end{figure}

Figure \ref{fig:Burgers_samples} provides a visual comparison between reference trajectory samples obtained by high-fidelity simulations of equation \ref{eq:Burgers} and trajectories generated by sampling the trained conditional generative model $p_{\theta}(\bm{u}|t,\bm{z})$. A more detailed comparison is provided in figure \ref{fig:Burgers_marginals} in terms of one-dimensional slices taken at four distinct time instances that were not used during model training. In both figures we observe a very good qualitative agreement between the reference and the predicted solutions, indicating that the conditional generative model is able to correctly capture the statistical structure of the system.  These results are indicative of the ability of the proposed method to approximate a non-trivial 128-dimensional distribution using only scattered measurements from 100 sample realizations of the system.

\section{Discussion}\label{sec:discussion}

We have presented a statistical inference framework for constructing scalable surrogate models for stochastic, high-dimensional and multi-fidelity systems. Leveraging recent advances in deep learning and stochastic variational inference, the proposed regularized inference procedure goes beyond mean-field and Gaussian approximations, it can accommodate the use of implicit models that are capable of approximating arbitrarily complex distributions, and is able to mitigate the issue of mode collapse that often hampers the performance of adversarial generative models. These elements enable the construction of conditional deep generative models that can be effectively trained on scattered and noisy input-output observations, and provide accurate predictions and robust uncertainty estimates. The latter, not only serves as a measure for a-posteriori error estimation, but it is also a key enabler of downstream tasks such as active learning \cite{cohn1996active} and Bayesian optimization \cite{shahriari2016taking}. Moreover,  the use of latent variables adds flexibility in learning from  data-sets that may be corrupted by complex noise processes, and offers a general platform for nonlinear dimensionality reduction. Taken all together, these developments aspire to provide a new set of probabilistic tools for expediting the analysis of stochastic systems, as well as act as unifying glue between experimental assays and computational modeling.

Our goal for this work is to present a new viewpoint on building surrogate models with a particular emphasis on the methodological foundations of the proposed algorithms. To this end, we confined the presentation to a diverse collection of canonical studies that were designed to highlight the broad applicability of the proposed tools, as well as to provide a test bed for systematic studies that elucidate their practical performance. In the process of gaining a deeper understanding of their advantages and limitations, future studies will focus on realistic large-scale applications in computational mechanics and beyond.

\section*{Acknowledgements}
This work received support from the US Department of Energy under the Advanced Scientific Computing Research program (grant DE-SC0019116) and the Defense Advanced Research Projects Agency under the Physics of Artificial Intelligence program.


\bibliographystyle{model1-num-names}
\bibliography{ref.bib}

\begin{thebibliography}{74}
\expandafter\ifx\csname natexlab\endcsname\relax\def\natexlab#1{#1}\fi
\providecommand{\bibinfo}[2]{#2}
\ifx\xfnm\relax \def\xfnm[#1]{\unskip,\space#1}\fi
\bibitem[{Forrester et~al.(2007)Forrester, S{\'o}bester, and
  Keane}]{forrester2007multi}
\bibinfo{author}{A.~I. Forrester}, \bibinfo{author}{A.~S{\'o}bester},
  \bibinfo{author}{A.~J. Keane},
\newblock \bibinfo{title}{Multi-fidelity optimization via surrogate modelling},
\newblock in: \bibinfo{booktitle}{Proceedings of the royal society of london a:
  mathematical, physical and engineering sciences}, volume
  \bibinfo{volume}{463}, \bibinfo{organization}{The Royal Society}, pp.
  \bibinfo{pages}{3251--3269}.
\bibitem[{Robinson et~al.(2008)Robinson, Eldred, Willcox, and
  Haimes}]{robinson2008surrogate}
\bibinfo{author}{T.~Robinson}, \bibinfo{author}{M.~Eldred},
  \bibinfo{author}{K.~Willcox}, \bibinfo{author}{R.~Haimes},
\newblock \bibinfo{title}{Surrogate-based optimization using multifidelity
  models with variable parameterization and corrected space mapping},
\newblock \bibinfo{journal}{AIAA Journal} \bibinfo{volume}{46}
  (\bibinfo{year}{2008}) \bibinfo{pages}{2814--2822}.
\bibitem[{Alexandrov et~al.(2001)Alexandrov, Lewis, Gumbert, Green, and
  Newman}]{alexandrov2001approximation}
\bibinfo{author}{N.~M. Alexandrov}, \bibinfo{author}{R.~M. Lewis},
  \bibinfo{author}{C.~R. Gumbert}, \bibinfo{author}{L.~L. Green},
  \bibinfo{author}{P.~A. Newman},
\newblock \bibinfo{title}{Approximation and model management in aerodynamic
  optimization with variable-fidelity models},
\newblock \bibinfo{journal}{Journal of Aircraft} \bibinfo{volume}{38}
  (\bibinfo{year}{2001}) \bibinfo{pages}{1093--1101}.
\bibitem[{Sun et~al.(2010)Sun, Li, Stone, and Li}]{sun2010two}
\bibinfo{author}{G.~Sun}, \bibinfo{author}{G.~Li}, \bibinfo{author}{M.~Stone},
  \bibinfo{author}{Q.~Li},
\newblock \bibinfo{title}{A two-stage multi-fidelity optimization procedure for
  honeycomb-type cellular materials},
\newblock \bibinfo{journal}{Computational Materials Science}
  \bibinfo{volume}{49} (\bibinfo{year}{2010}) \bibinfo{pages}{500--511}.
\bibitem[{Sun et~al.(2011)Sun, Li, Zhou, Xu, Yang, and Li}]{sun2011multi}
\bibinfo{author}{G.~Sun}, \bibinfo{author}{G.~Li}, \bibinfo{author}{S.~Zhou},
  \bibinfo{author}{W.~Xu}, \bibinfo{author}{X.~Yang}, \bibinfo{author}{Q.~Li},
\newblock \bibinfo{title}{Multi-fidelity optimization for sheet metal forming
  process},
\newblock \bibinfo{journal}{Structural and Multidisciplinary Optimization}
  \bibinfo{volume}{44} (\bibinfo{year}{2011}) \bibinfo{pages}{111--124}.
\bibitem[{Celik et~al.(2010)Celik, Lee, Vasudevan, and Son}]{celik2010dddas}
\bibinfo{author}{N.~Celik}, \bibinfo{author}{S.~Lee},
  \bibinfo{author}{K.~Vasudevan}, \bibinfo{author}{Y.-J. Son},
\newblock \bibinfo{title}{Dddas-based multi-fidelity simulation framework for
  supply chain systems},
\newblock \bibinfo{journal}{IIE Transactions} \bibinfo{volume}{42}
  (\bibinfo{year}{2010}) \bibinfo{pages}{325--341}.
\bibitem[{Perdikaris and Karniadakis(2016)}]{perdikaris2016model}
\bibinfo{author}{P.~Perdikaris}, \bibinfo{author}{G.~E. Karniadakis},
\newblock \bibinfo{title}{Model inversion via multi-fidelity {B}ayesian
  optimization: a new paradigm for parameter estimation in haemodynamics, and
  beyond},
\newblock \bibinfo{journal}{Journal of The Royal Society Interface}
  \bibinfo{volume}{13} (\bibinfo{year}{2016}) \bibinfo{pages}{20151107}.
\bibitem[{Perdikaris(2015)}]{perdikaris2015data}
\bibinfo{author}{P.~Perdikaris}, \bibinfo{title}{Data-Driven Parallel
  Scientific Computing: Multi-Fidelity Information Fusion Algorithms and
  Applications to Physical and Biological Systems}, Ph.D. thesis, Brown
  University, \bibinfo{year}{2015}.
\bibitem[{Perdikaris and Karniadakis(2015)}]{perdikaris2015calibration}
\bibinfo{author}{P.~Perdikaris}, \bibinfo{author}{G.~Karniadakis},
\newblock \bibinfo{title}{Calibration of blood flow in simulations via
  multi-fidelity {B}ayesian optimization},
\newblock in: \bibinfo{booktitle}{APS Meeting Abstracts}.
\bibitem[{Eldred and Burkardt(2009)}]{eldred2009comparison}
\bibinfo{author}{M.~Eldred}, \bibinfo{author}{J.~Burkardt},
\newblock \bibinfo{title}{Comparison of non-intrusive polynomial chaos and
  stochastic collocation methods for uncertainty quantification},
\newblock in: \bibinfo{booktitle}{47th {AIAA} Aerospace Sciences Meeting
  including The New Horizons Forum and Aerospace Exposition}, p.
  \bibinfo{pages}{976}.
\bibitem[{Ng and Eldred(2012)}]{ng2012multifidelity}
\bibinfo{author}{L.~W.-T. Ng}, \bibinfo{author}{M.~Eldred},
\newblock \bibinfo{title}{Multifidelity uncertainty quantification using
  non-intrusive polynomial chaos and stochastic collocation},
\newblock in: \bibinfo{booktitle}{53rd {AIAA}/ASME/ASCE/AHS/ASC Structures,
  Structural Dynamics and Materials Conference 20th {AIAA}/ASME/AHS Adaptive
  Structures Conference 14th {AIAA}}, p. \bibinfo{pages}{1852}.
\bibitem[{Padron et~al.(2014)Padron, Alonso, Palacios, Barone, and
  Eldred}]{padron2014multi}
\bibinfo{author}{A.~S. Padron}, \bibinfo{author}{J.~J. Alonso},
  \bibinfo{author}{F.~Palacios}, \bibinfo{author}{M.~F. Barone},
  \bibinfo{author}{M.~S. Eldred},
\newblock \bibinfo{title}{Multi-fidelity uncertainty quantification:
  application to a vertical axis wind turbine under an extreme gust},
\newblock in: \bibinfo{booktitle}{15th {AIAA}/ISSMO Multidisciplinary Analysis
  and Optimization Conference}, p. \bibinfo{pages}{3013}.
\bibitem[{Biehler et~al.(2015)Biehler, Gee, and Wall}]{biehler2015towards}
\bibinfo{author}{J.~Biehler}, \bibinfo{author}{M.~W. Gee},
  \bibinfo{author}{W.~A. Wall},
\newblock \bibinfo{title}{Towards efficient uncertainty quantification in
  complex and large-scale biomechanical problems based on a {B}ayesian
  multi-fidelity scheme},
\newblock \bibinfo{journal}{Biomechanics and modeling in mechanobiology}
  \bibinfo{volume}{14} (\bibinfo{year}{2015}) \bibinfo{pages}{489--513}.
\bibitem[{Peherstorfer et~al.(2016{\natexlab{a}})Peherstorfer, Willcox, and
  Gunzburger}]{peherstorfer2016optimal}
\bibinfo{author}{B.~Peherstorfer}, \bibinfo{author}{K.~Willcox},
  \bibinfo{author}{M.~Gunzburger},
\newblock \bibinfo{title}{Optimal model management for multifidelity monte
  carlo estimation},
\newblock \bibinfo{journal}{SIAM Journal on Scientific Computing}
  \bibinfo{volume}{38} (\bibinfo{year}{2016}{\natexlab{a}})
  \bibinfo{pages}{A3163--A3194}.
\bibitem[{Peherstorfer et~al.(2016{\natexlab{b}})Peherstorfer, Cui, Marzouk,
  and Willcox}]{peherstorfer2016multifidelity}
\bibinfo{author}{B.~Peherstorfer}, \bibinfo{author}{T.~Cui},
  \bibinfo{author}{Y.~Marzouk}, \bibinfo{author}{K.~Willcox},
\newblock \bibinfo{title}{Multifidelity importance sampling},
\newblock \bibinfo{journal}{Computer Methods in Applied Mechanics and
  Engineering} \bibinfo{volume}{300} (\bibinfo{year}{2016}{\natexlab{b}})
  \bibinfo{pages}{490--509}.
\bibitem[{Peherstorfer et~al.(2016{\natexlab{c}})Peherstorfer, Willcox, and
  Gunzburger}]{peherstorfer2016survey}
\bibinfo{author}{B.~Peherstorfer}, \bibinfo{author}{K.~Willcox},
  \bibinfo{author}{M.~Gunzburger},
\newblock \bibinfo{title}{Survey of multifidelity methods in uncertainty
  propagation, inference, and optimization},
\newblock \bibinfo{journal}{Preprint}  (\bibinfo{year}{2016}{\natexlab{c}})
  \bibinfo{pages}{1--57}.
\bibitem[{Narayan et~al.(2014)Narayan, Gittelson, and
  Xiu}]{narayan2014stochastic}
\bibinfo{author}{A.~Narayan}, \bibinfo{author}{C.~Gittelson},
  \bibinfo{author}{D.~Xiu},
\newblock \bibinfo{title}{A stochastic collocation algorithm with multifidelity
  models},
\newblock \bibinfo{journal}{SIAM Journal on Scientific Computing}
  \bibinfo{volume}{36} (\bibinfo{year}{2014}) \bibinfo{pages}{A495--A521}.
\bibitem[{Zhu et~al.(2014)Zhu, Narayan, and Xiu}]{zhu2014computational}
\bibinfo{author}{X.~Zhu}, \bibinfo{author}{A.~Narayan},
  \bibinfo{author}{D.~Xiu},
\newblock \bibinfo{title}{Computational aspects of stochastic collocation with
  multifidelity models},
\newblock \bibinfo{journal}{SIAM/ASA Journal on Uncertainty Quantification}
  \bibinfo{volume}{2} (\bibinfo{year}{2014}) \bibinfo{pages}{444--463}.
\bibitem[{Bilionis et~al.(2013)Bilionis, Zabaras, Konomi, and
  Lin}]{bilionis2013multi}
\bibinfo{author}{I.~Bilionis}, \bibinfo{author}{N.~Zabaras},
  \bibinfo{author}{B.~A. Konomi}, \bibinfo{author}{G.~Lin},
\newblock \bibinfo{title}{Multi-output separable {G}aussian process: Towards an
  efficient, fully {B}ayesian paradigm for uncertainty quantification},
\newblock \bibinfo{journal}{Journal of Computational Physics}
  \bibinfo{volume}{241} (\bibinfo{year}{2013}) \bibinfo{pages}{212--239}.
\bibitem[{Parussini et~al.(2017)Parussini, Venturi, Perdikaris, and
  Karniadakis}]{parussini2017multi}
\bibinfo{author}{L.~Parussini}, \bibinfo{author}{D.~Venturi},
  \bibinfo{author}{P.~Perdikaris}, \bibinfo{author}{G.~Karniadakis},
\newblock \bibinfo{title}{Multi-fidelity {G}aussian process regression for
  prediction of random fields},
\newblock \bibinfo{journal}{J. Comput. Phys.} \bibinfo{volume}{336}
  (\bibinfo{year}{2017}) \bibinfo{pages}{36 -- 50}.
\bibitem[{Perdikaris et~al.(2016)Perdikaris, Venturi, and
  Karniadakis}]{perdikaris2016multifidelity}
\bibinfo{author}{P.~Perdikaris}, \bibinfo{author}{D.~Venturi},
  \bibinfo{author}{G.~E. Karniadakis},
\newblock \bibinfo{title}{Multifidelity information fusion algorithms for
  high-dimensional systems and massive data sets},
\newblock \bibinfo{journal}{SIAM J. Sci. Comput.} \bibinfo{volume}{38}
  (\bibinfo{year}{2016}) \bibinfo{pages}{B521--B538}.
\bibitem[{Rasmussen(2004)}]{rasmussen2004gaussian}
\bibinfo{author}{C.~E. Rasmussen},
\newblock \bibinfo{title}{Gaussian processes in machine learning},
\newblock in: \bibinfo{booktitle}{Advanced lectures on machine learning},
  \bibinfo{publisher}{Springer}, \bibinfo{year}{2004}, pp.
  \bibinfo{pages}{63--71}.
\bibitem[{Kingma and Welling(2013)}]{kingma2013auto}
\bibinfo{author}{D.~P. Kingma}, \bibinfo{author}{M.~Welling},
\newblock \bibinfo{title}{Auto-encoding variational {B}ayes},
\newblock \bibinfo{journal}{arXiv preprint arXiv:1312.6114}
  (\bibinfo{year}{2013}).
\bibitem[{Sohn et~al.(2015)Sohn, Lee, and Yan}]{sohn2015learning}
\bibinfo{author}{K.~Sohn}, \bibinfo{author}{H.~Lee}, \bibinfo{author}{X.~Yan},
\newblock \bibinfo{title}{Learning structured output representation using deep
  conditional generative models},
\newblock in: \bibinfo{booktitle}{Advances in Neural Information Processing
  Systems}, pp. \bibinfo{pages}{3483--3491}.
\bibitem[{Vincent et~al.(2008)Vincent, Larochelle, Bengio, and
  Manzagol}]{vincent2008extracting}
\bibinfo{author}{P.~Vincent}, \bibinfo{author}{H.~Larochelle},
  \bibinfo{author}{Y.~Bengio}, \bibinfo{author}{P.-A. Manzagol},
\newblock \bibinfo{title}{Extracting and composing robust features with
  denoising autoencoders},
\newblock in: \bibinfo{booktitle}{Proceedings of the 25th international
  conference on Machine learning}, \bibinfo{organization}{ACM}, pp.
  \bibinfo{pages}{1096--1103}.
\bibitem[{Vincent et~al.(2010)Vincent, Larochelle, Lajoie, Bengio, and
  Manzagol}]{vincent2010stacked}
\bibinfo{author}{P.~Vincent}, \bibinfo{author}{H.~Larochelle},
  \bibinfo{author}{I.~Lajoie}, \bibinfo{author}{Y.~Bengio},
  \bibinfo{author}{P.-A. Manzagol},
\newblock \bibinfo{title}{Stacked denoising autoencoders: Learning useful
  representations in a deep network with a local denoising criterion},
\newblock \bibinfo{journal}{Journal of Machine Learning Research}
  \bibinfo{volume}{11} (\bibinfo{year}{2010}) \bibinfo{pages}{3371--3408}.
\bibitem[{G{\'o}mez-Bombarelli(2016)}]{gomez2016design}
\bibinfo{author}{R.~{\em e. al.}. G{\'o}mez-Bombarelli},
\newblock \bibinfo{title}{Design of efficient molecular organic light-emitting
  diodes by a high-throughput virtual screening and experimental approach},
\newblock \bibinfo{journal}{Nature materials} \bibinfo{volume}{15}
  (\bibinfo{year}{2016}) \bibinfo{pages}{1120--1127}.
\bibitem[{G{\'o}mez-Bombarelli et~al.(2018)G{\'o}mez-Bombarelli, Wei, Duvenaud,
  Hern{\'a}ndez-Lobato, S{\'a}nchez-Lengeling, Sheberla, Aguilera-Iparraguirre,
  Hirzel, Adams, and Aspuru-Guzik}]{gomez2018automatic}
\bibinfo{author}{R.~G{\'o}mez-Bombarelli}, \bibinfo{author}{J.~N. Wei},
  \bibinfo{author}{D.~Duvenaud}, \bibinfo{author}{J.~M. Hern{\'a}ndez-Lobato},
  \bibinfo{author}{B.~S{\'a}nchez-Lengeling}, \bibinfo{author}{D.~Sheberla},
  \bibinfo{author}{J.~Aguilera-Iparraguirre}, \bibinfo{author}{T.~D. Hirzel},
  \bibinfo{author}{R.~P. Adams}, \bibinfo{author}{A.~Aspuru-Guzik},
\newblock \bibinfo{title}{Automatic chemical design using a data-driven
  continuous representation of molecules},
\newblock \bibinfo{journal}{ACS Central Science} \bibinfo{volume}{4}
  (\bibinfo{year}{2018}) \bibinfo{pages}{268--276}.
\bibitem[{Ravanbakhsh et~al.(2017)Ravanbakhsh, Lanusse, Mandelbaum, Schneider,
  and Poczos}]{ravanbakhsh2017enabling}
\bibinfo{author}{S.~Ravanbakhsh}, \bibinfo{author}{F.~Lanusse},
  \bibinfo{author}{R.~Mandelbaum}, \bibinfo{author}{J.~G. Schneider},
  \bibinfo{author}{B.~Poczos},
\newblock \bibinfo{title}{Enabling dark energy science with deep generative
  models of galaxy images.},
\newblock in: \bibinfo{booktitle}{AAAI}, pp. \bibinfo{pages}{1488--1494}.
\bibitem[{Lopez et~al.(2017)Lopez, Regier, Cole, Jordan, and
  Yosef}]{lopez2017deep}
\bibinfo{author}{R.~Lopez}, \bibinfo{author}{J.~Regier},
  \bibinfo{author}{M.~Cole}, \bibinfo{author}{M.~Jordan},
  \bibinfo{author}{N.~Yosef},
\newblock \bibinfo{title}{A deep generative model for single-cell {RNA}
  sequencing with application to detecting differentially expressed genes},
\newblock \bibinfo{journal}{arXiv preprint arXiv:1710.05086}
  (\bibinfo{year}{2017}).
\bibitem[{Way and Greene(2017)}]{way2017extracting}
\bibinfo{author}{G.~P. Way}, \bibinfo{author}{C.~S. Greene},
\newblock \bibinfo{title}{Extracting a biologically relevant latent space from
  cancer transcriptomes with variational autoencoders},
\newblock \bibinfo{journal}{bioRxiv}  (\bibinfo{year}{2017})
  \bibinfo{pages}{174474}.
\bibitem[{Bousquet et~al.(2017)Bousquet, Gelly, Tolstikhin, Simon-Gabriel, and
  Schoelkopf}]{bousquet2017optimal}
\bibinfo{author}{O.~Bousquet}, \bibinfo{author}{S.~Gelly},
  \bibinfo{author}{I.~Tolstikhin}, \bibinfo{author}{C.-J. Simon-Gabriel},
  \bibinfo{author}{B.~Schoelkopf},
\newblock \bibinfo{title}{From optimal transport to generative modeling: the
  {VEGAN} cookbook},
\newblock \bibinfo{journal}{arXiv preprint arXiv:1705.07642}
  (\bibinfo{year}{2017}).
\bibitem[{Pu et~al.(2017)Pu, Chen, Dai, Wang, Li, and Carin}]{pu2017symmetric}
\bibinfo{author}{Y.~Pu}, \bibinfo{author}{L.~Chen}, \bibinfo{author}{S.~Dai},
  \bibinfo{author}{W.~Wang}, \bibinfo{author}{C.~Li},
  \bibinfo{author}{L.~Carin},
\newblock \bibinfo{title}{Symmetric variational autoencoder and connections to
  adversarial learning},
\newblock \bibinfo{journal}{arXiv preprint arXiv:1709.01846}
  (\bibinfo{year}{2017}).
\bibitem[{Rosca et~al.(2018)Rosca, Lakshminarayanan, and
  Mohamed}]{rosca2018distribution}
\bibinfo{author}{M.~Rosca}, \bibinfo{author}{B.~Lakshminarayanan},
  \bibinfo{author}{S.~Mohamed},
\newblock \bibinfo{title}{Distribution matching in variational inference},
\newblock \bibinfo{journal}{arXiv preprint arXiv:1802.06847}
  (\bibinfo{year}{2018}).
\bibitem[{Zheng et~al.(2018)Zheng, Yao, Zhang, and
  Tsang}]{zheng2018degeneration}
\bibinfo{author}{H.~Zheng}, \bibinfo{author}{J.~Yao},
  \bibinfo{author}{Y.~Zhang}, \bibinfo{author}{I.~W. Tsang},
\newblock \bibinfo{title}{Degeneration in {VAE}: in the light of fisher
  information loss},
\newblock \bibinfo{journal}{arXiv preprint arXiv:1802.06677}
  (\bibinfo{year}{2018}).
\bibitem[{Kingma et~al.(2016)Kingma, Salimans, Jozefowicz, Chen, Sutskever, and
  Welling}]{kingma2016improved}
\bibinfo{author}{D.~P. Kingma}, \bibinfo{author}{T.~Salimans},
  \bibinfo{author}{R.~Jozefowicz}, \bibinfo{author}{X.~Chen},
  \bibinfo{author}{I.~Sutskever}, \bibinfo{author}{M.~Welling},
\newblock \bibinfo{title}{Improved variational inference with inverse
  autoregressive flow},
\newblock in: \bibinfo{booktitle}{Advances in Neural Information Processing
  Systems}, pp. \bibinfo{pages}{4743--4751}.
\bibitem[{Rezende and Mohamed(2015)}]{rezende2015variational}
\bibinfo{author}{D.~J. Rezende}, \bibinfo{author}{S.~Mohamed},
\newblock \bibinfo{title}{Variational inference with normalizing flows},
\newblock \bibinfo{journal}{arXiv preprint arXiv:1505.05770}
  (\bibinfo{year}{2015}).
\bibitem[{Higgins et~al.(2016)Higgins, Matthey, Pal, Burgess, Glorot,
  Botvinick, Mohamed, and Lerchner}]{higgins2016beta}
\bibinfo{author}{I.~Higgins}, \bibinfo{author}{L.~Matthey},
  \bibinfo{author}{A.~Pal}, \bibinfo{author}{C.~Burgess},
  \bibinfo{author}{X.~Glorot}, \bibinfo{author}{M.~Botvinick},
  \bibinfo{author}{S.~Mohamed}, \bibinfo{author}{A.~Lerchner},
\newblock \bibinfo{title}{beta-{VAE}: Learning basic visual concepts with a
  constrained variational framework}  (\bibinfo{year}{2016}).
\bibitem[{Zhao et~al.(2017)Zhao, Song, and Ermon}]{zhao2017infovae}
\bibinfo{author}{S.~Zhao}, \bibinfo{author}{J.~Song},
  \bibinfo{author}{S.~Ermon},
\newblock \bibinfo{title}{Info{VAE}: Information maximizing variational
  autoencoders},
\newblock \bibinfo{journal}{arXiv preprint arXiv:1706.02262}
  (\bibinfo{year}{2017}).
\bibitem[{Chen et~al.(2018)Chen, Li, Grosse, and Duvenaud}]{chen2018isolating}
\bibinfo{author}{T.~Q. Chen}, \bibinfo{author}{X.~Li},
  \bibinfo{author}{R.~Grosse}, \bibinfo{author}{D.~Duvenaud},
\newblock \bibinfo{title}{Isolating sources of disentanglement in variational
  autoencoders},
\newblock \bibinfo{journal}{arXiv preprint arXiv:1802.04942}
  (\bibinfo{year}{2018}).
\bibitem[{Burda et~al.(2015)Burda, Grosse, and
  Salakhutdinov}]{burda2015importance}
\bibinfo{author}{Y.~Burda}, \bibinfo{author}{R.~Grosse},
  \bibinfo{author}{R.~Salakhutdinov},
\newblock \bibinfo{title}{Importance weighted autoencoders},
\newblock \bibinfo{journal}{arXiv preprint arXiv:1509.00519}
  (\bibinfo{year}{2015}).
\bibitem[{Klys et~al.(2018)Klys, Bettencourt, and Duvenaud}]{klys2018joint}
\bibinfo{author}{J.~Klys}, \bibinfo{author}{J.~Bettencourt},
  \bibinfo{author}{D.~Duvenaud},
\newblock \bibinfo{title}{Joint importance sampling for variational inference}
  (\bibinfo{year}{2018}).
\bibitem[{Genevay et~al.(2017)Genevay, Peyr{\'e}, and Cuturi}]{genevay2017gan}
\bibinfo{author}{A.~Genevay}, \bibinfo{author}{G.~Peyr{\'e}},
  \bibinfo{author}{M.~Cuturi},
\newblock \bibinfo{title}{{GAN} and {VAE} from an optimal transport point of
  view},
\newblock \bibinfo{journal}{arXiv preprint arXiv:1706.01807}
  (\bibinfo{year}{2017}).
\bibitem[{Villani(2008)}]{villani2008optimal}
\bibinfo{author}{C.~Villani}, \bibinfo{title}{Optimal transport: old and new},
  volume \bibinfo{volume}{338}, \bibinfo{publisher}{Springer Science \&
  Business Media}, \bibinfo{year}{2008}.
\bibitem[{El~Moselhy and Marzouk(2012)}]{el2012bayesian}
\bibinfo{author}{T.~A. El~Moselhy}, \bibinfo{author}{Y.~M. Marzouk},
\newblock \bibinfo{title}{{B}ayesian inference with optimal maps},
\newblock \bibinfo{journal}{Journal of Computational Physics}
  \bibinfo{volume}{231} (\bibinfo{year}{2012}) \bibinfo{pages}{7815--7850}.
\bibitem[{van~den Oord et~al.(2016)van~den Oord, Kalchbrenner, Espeholt,
  Vinyals, Graves et~al.}]{van2016conditional}
\bibinfo{author}{A.~van~den Oord}, \bibinfo{author}{N.~Kalchbrenner},
  \bibinfo{author}{L.~Espeholt}, \bibinfo{author}{O.~Vinyals},
  \bibinfo{author}{A.~Graves}, et~al.,
\newblock \bibinfo{title}{Conditional image generation with pixelcnn decoders},
\newblock in: \bibinfo{booktitle}{Advances in Neural Information Processing
  Systems}, pp. \bibinfo{pages}{4790--4798}.
\bibitem[{Liu and Wang(2016)}]{liu2016stein}
\bibinfo{author}{Q.~Liu}, \bibinfo{author}{D.~Wang},
\newblock \bibinfo{title}{Stein variational gradient descent: A general purpose
  bayesian inference algorithm},
\newblock in: \bibinfo{booktitle}{Advances In Neural Information Processing
  Systems}, pp. \bibinfo{pages}{2378--2386}.
\bibitem[{Mescheder et~al.(2017)Mescheder, Nowozin, and
  Geiger}]{mescheder2017adversarial}
\bibinfo{author}{L.~Mescheder}, \bibinfo{author}{S.~Nowozin},
  \bibinfo{author}{A.~Geiger},
\newblock \bibinfo{title}{Adversarial variational bayes: Unifying variational
  autoencoders and generative adversarial networks},
\newblock \bibinfo{journal}{arXiv preprint arXiv:1701.04722}
  (\bibinfo{year}{2017}).
\bibitem[{Makhzani et~al.(2015)Makhzani, Shlens, Jaitly, Goodfellow, and
  Frey}]{makhzani2015adversarial}
\bibinfo{author}{A.~Makhzani}, \bibinfo{author}{J.~Shlens},
  \bibinfo{author}{N.~Jaitly}, \bibinfo{author}{I.~Goodfellow},
  \bibinfo{author}{B.~Frey},
\newblock \bibinfo{title}{Adversarial autoencoders},
\newblock \bibinfo{journal}{arXiv preprint arXiv:1511.05644}
  (\bibinfo{year}{2015}).
\bibitem[{Tolstikhin et~al.(2017)Tolstikhin, Bousquet, Gelly, and
  Schoelkopf}]{tolstikhin2017wasserstein}
\bibinfo{author}{I.~Tolstikhin}, \bibinfo{author}{O.~Bousquet},
  \bibinfo{author}{S.~Gelly}, \bibinfo{author}{B.~Schoelkopf},
\newblock \bibinfo{title}{Wasserstein auto-encoders},
\newblock \bibinfo{journal}{arXiv preprint arXiv:1711.01558}
  (\bibinfo{year}{2017}).
\bibitem[{Titsias(2017)}]{titsias2017learning}
\bibinfo{author}{M.~K. Titsias},
\newblock \bibinfo{title}{Learning model reparametrizations: Implicit
  variational inference by fitting mcmc distributions},
\newblock \bibinfo{journal}{arXiv preprint arXiv:1708.01529}
  (\bibinfo{year}{2017}).
\bibitem[{Blei et~al.(2017)Blei, Kucukelbir, and
  McAuliffe}]{blei2017variational}
\bibinfo{author}{D.~M. Blei}, \bibinfo{author}{A.~Kucukelbir},
  \bibinfo{author}{J.~D. McAuliffe},
\newblock \bibinfo{title}{Variational inference: A review for statisticians},
\newblock \bibinfo{journal}{Journal of the American Statistical Association}
  \bibinfo{volume}{112} (\bibinfo{year}{2017}) \bibinfo{pages}{859--877}.
\bibitem[{Wainwright et~al.(2008)Wainwright, Jordan
  et~al.}]{wainwright2008graphical}
\bibinfo{author}{M.~J. Wainwright}, \bibinfo{author}{M.~I. Jordan}, et~al.,
\newblock \bibinfo{title}{Graphical models, exponential families, and
  variational inference},
\newblock \bibinfo{journal}{Foundations and Trends{\textregistered} in Machine
  Learning} \bibinfo{volume}{1} (\bibinfo{year}{2008}) \bibinfo{pages}{1--305}.
\bibitem[{Goodfellow et~al.(2014)Goodfellow, Pouget-Abadie, Mirza, Xu,
  Warde-Farley, Ozair, Courville, and Bengio}]{goodfellow2014generative}
\bibinfo{author}{I.~Goodfellow}, \bibinfo{author}{J.~Pouget-Abadie},
  \bibinfo{author}{M.~Mirza}, \bibinfo{author}{B.~Xu},
  \bibinfo{author}{D.~Warde-Farley}, \bibinfo{author}{S.~Ozair},
  \bibinfo{author}{A.~Courville}, \bibinfo{author}{Y.~Bengio},
\newblock \bibinfo{title}{Generative adversarial nets},
\newblock in: \bibinfo{booktitle}{Advances in neural information processing
  systems}, pp. \bibinfo{pages}{2672--2680}.
\bibitem[{Li et~al.(2018)Li, Li, Wang, and Carin}]{li2018learning}
\bibinfo{author}{C.~Li}, \bibinfo{author}{J.~Li}, \bibinfo{author}{G.~Wang},
  \bibinfo{author}{L.~Carin},
\newblock \bibinfo{title}{Learning to sample with adversarially learned
  likelihood-ratio}  (\bibinfo{year}{2018}).
\bibitem[{Salimans et~al.(2016)Salimans, Goodfellow, Zaremba, Cheung, Radford,
  and Chen}]{salimans2016improved}
\bibinfo{author}{T.~Salimans}, \bibinfo{author}{I.~Goodfellow},
  \bibinfo{author}{W.~Zaremba}, \bibinfo{author}{V.~Cheung},
  \bibinfo{author}{A.~Radford}, \bibinfo{author}{X.~Chen},
\newblock \bibinfo{title}{Improved techniques for training gans},
\newblock in: \bibinfo{booktitle}{Advances in Neural Information Processing
  Systems}, pp. \bibinfo{pages}{2234--2242}.
\bibitem[{Akaike(1998)}]{akaike1998information}
\bibinfo{author}{H.~Akaike},
\newblock \bibinfo{title}{Information theory and an extension of the maximum
  likelihood principle},
\newblock in: \bibinfo{booktitle}{Selected papers of hirotugu akaike},
  \bibinfo{publisher}{Springer}, \bibinfo{year}{1998}, pp.
  \bibinfo{pages}{199--213}.
\bibitem[{Friedman et~al.(2001)Friedman, Hastie, and
  Tibshirani}]{friedman2001elements}
\bibinfo{author}{J.~Friedman}, \bibinfo{author}{T.~Hastie},
  \bibinfo{author}{R.~Tibshirani}, \bibinfo{title}{The elements of statistical
  learning}, volume~\bibinfo{volume}{1}, \bibinfo{publisher}{Springer series in
  statistics New York, NY, USA:}, \bibinfo{year}{2001}.
\bibitem[{Kingma and Ba(2014)}]{kingma2014adam}
\bibinfo{author}{D.~P. Kingma}, \bibinfo{author}{J.~Ba},
\newblock \bibinfo{title}{Adam: A method for stochastic optimization},
\newblock \bibinfo{journal}{arXiv preprint arXiv:1412.6980}
  (\bibinfo{year}{2014}).
\bibitem[{Abadi et~al.(2016)Abadi, Barham, Chen, Chen, Davis, Dean, Devin,
  Ghemawat, Irving, Isard et~al.}]{abadi2016tensorflow}
\bibinfo{author}{M.~Abadi}, \bibinfo{author}{P.~Barham},
  \bibinfo{author}{J.~Chen}, \bibinfo{author}{Z.~Chen},
  \bibinfo{author}{A.~Davis}, \bibinfo{author}{J.~Dean},
  \bibinfo{author}{M.~Devin}, \bibinfo{author}{S.~Ghemawat},
  \bibinfo{author}{G.~Irving}, \bibinfo{author}{M.~Isard}, et~al.,
\newblock \bibinfo{title}{Tensorflow: A system for large-scale machine
  learning.},
\newblock in: \bibinfo{booktitle}{OSDI}, volume~\bibinfo{volume}{16}, pp.
  \bibinfo{pages}{265--283}.
\bibitem[{Goodfellow et~al.(2016)Goodfellow, Bengio, Courville, and
  Bengio}]{goodfellow2016deep}
\bibinfo{author}{I.~Goodfellow}, \bibinfo{author}{Y.~Bengio},
  \bibinfo{author}{A.~Courville}, \bibinfo{author}{Y.~Bengio},
  \bibinfo{title}{Deep learning}, volume~\bibinfo{volume}{1},
  \bibinfo{publisher}{MIT press Cambridge}, \bibinfo{year}{2016}.
\bibitem[{Neal(2012)}]{neal2012bayesian}
\bibinfo{author}{R.~M. Neal}, \bibinfo{title}{Bayesian learning for neural
  networks}, volume \bibinfo{volume}{118}, \bibinfo{publisher}{Springer Science
  \& Business Media}, \bibinfo{year}{2012}.
\bibitem[{Kennedy and O'Hagan(2000)}]{kennedy2000predicting}
\bibinfo{author}{M.~C. Kennedy}, \bibinfo{author}{A.~O'Hagan},
\newblock \bibinfo{title}{Predicting the output from a complex computer code
  when fast approximations are available},
\newblock \bibinfo{journal}{Biometrika} \bibinfo{volume}{87}
  (\bibinfo{year}{2000}) \bibinfo{pages}{1--13}.
\bibitem[{Perdikaris et~al.(2017)Perdikaris, Raissi, Damianou, Lawrence, and
  Karniadakis}]{perdikaris2017nonlinear}
\bibinfo{author}{P.~Perdikaris}, \bibinfo{author}{M.~Raissi},
  \bibinfo{author}{A.~Damianou}, \bibinfo{author}{N.~Lawrence},
  \bibinfo{author}{G.~Karniadakis},
\newblock \bibinfo{title}{Nonlinear information fusion algorithms for
  data-efficient multi-fidelity modelling},
\newblock in: \bibinfo{booktitle}{Proc. R. Soc. A}, volume
  \bibinfo{volume}{473}, \bibinfo{organization}{The Royal Society}, p.
  \bibinfo{pages}{20160751}.
\bibitem[{Burgers(1948)}]{burgers1948mathematical}
\bibinfo{author}{J.~M. Burgers},
\newblock \bibinfo{title}{A mathematical model illustrating the theory of
  turbulence},
\newblock in: \bibinfo{booktitle}{Advances in applied mechanics},
  volume~\bibinfo{volume}{1}, \bibinfo{publisher}{Elsevier},
  \bibinfo{year}{1948}, pp. \bibinfo{pages}{171--199}.
\bibitem[{Kassam and Trefethen(2005)}]{kassam2005fourth}
\bibinfo{author}{A.-K. Kassam}, \bibinfo{author}{L.~N. Trefethen},
\newblock \bibinfo{title}{Fourth-order time-stepping for stiff pdes},
\newblock \bibinfo{journal}{SIAM Journal on Scientific Computing}
  \bibinfo{volume}{26} (\bibinfo{year}{2005}) \bibinfo{pages}{1214--1233}.
\bibitem[{Krizhevsky et~al.(2012)Krizhevsky, Sutskever, and
  Hinton}]{krizhevsky2012imagenet}
\bibinfo{author}{A.~Krizhevsky}, \bibinfo{author}{I.~Sutskever},
  \bibinfo{author}{G.~E. Hinton},
\newblock \bibinfo{title}{Imagenet classification with deep convolutional
  neural networks},
\newblock in: \bibinfo{booktitle}{Advances in neural information processing
  systems}, pp. \bibinfo{pages}{1097--1105}.
\bibitem[{LeCun et~al.(2015)LeCun, Bengio, and Hinton}]{lecun2015deep}
\bibinfo{author}{Y.~LeCun}, \bibinfo{author}{Y.~Bengio},
  \bibinfo{author}{G.~Hinton},
\newblock \bibinfo{title}{Deep learning},
\newblock \bibinfo{journal}{Nature} \bibinfo{volume}{521}
  (\bibinfo{year}{2015}) \bibinfo{pages}{436--444}.
\bibitem[{Mallat(2016)}]{mallat2016understanding}
\bibinfo{author}{S.~Mallat},
\newblock \bibinfo{title}{Understanding deep convolutional networks},
\newblock \bibinfo{journal}{Phil. Trans. R. Soc. A} \bibinfo{volume}{374}
  (\bibinfo{year}{2016}) \bibinfo{pages}{20150203}.
\bibitem[{Ioffe and Szegedy(2015)}]{ioffe2015batch}
\bibinfo{author}{S.~Ioffe}, \bibinfo{author}{C.~Szegedy},
\newblock \bibinfo{title}{Batch normalization: Accelerating deep network
  training by reducing internal covariate shift},
\newblock \bibinfo{journal}{arXiv preprint arXiv:1502.03167}
  (\bibinfo{year}{2015}).
\bibitem[{Cohn et~al.(1996)Cohn, Ghahramani, and Jordan}]{cohn1996active}
\bibinfo{author}{D.~A. Cohn}, \bibinfo{author}{Z.~Ghahramani},
  \bibinfo{author}{M.~I. Jordan},
\newblock \bibinfo{title}{Active learning with statistical models},
\newblock \bibinfo{journal}{Journal of artificial intelligence research}
  (\bibinfo{year}{1996}).
\bibitem[{Shahriari et~al.(2016)Shahriari, Swersky, Wang, Adams, and
  De~Freitas}]{shahriari2016taking}
\bibinfo{author}{B.~Shahriari}, \bibinfo{author}{K.~Swersky},
  \bibinfo{author}{Z.~Wang}, \bibinfo{author}{R.~P. Adams},
  \bibinfo{author}{N.~De~Freitas},
\newblock \bibinfo{title}{Taking the human out of the loop: A review of
  {B}ayesian optimization},
\newblock \bibinfo{journal}{Proceedings of the IEEE} \bibinfo{volume}{104}
  (\bibinfo{year}{2016}) \bibinfo{pages}{148--175}.
\bibitem[{Glorot and Bengio(2010)}]{glorot2010understanding}
\bibinfo{author}{X.~Glorot}, \bibinfo{author}{Y.~Bengio},
\newblock \bibinfo{title}{Understanding the difficulty of training deep
  feedforward neural networks},
\newblock in: \bibinfo{booktitle}{Proceedings of the thirteenth international
  conference on artificial intelligence and statistics}, pp.
  \bibinfo{pages}{249--256}.
\bibitem[{Yang and Perdikaris(2018)}]{yang2018adversarial}
\bibinfo{author}{Y.~Yang}, \bibinfo{author}{P.~Perdikaris},
\newblock \bibinfo{title}{Adversarial uncertainty quantification in
  physics-informed neural networks},
\newblock \bibinfo{journal}{arXiv preprint arXiv:1811.04026}
  (\bibinfo{year}{2018}).

\end{thebibliography}

\appendix

\section{Sensitivity studies}
\label{sec:appendix}

\begin{figure}
\centering
\includegraphics[width=\textwidth]{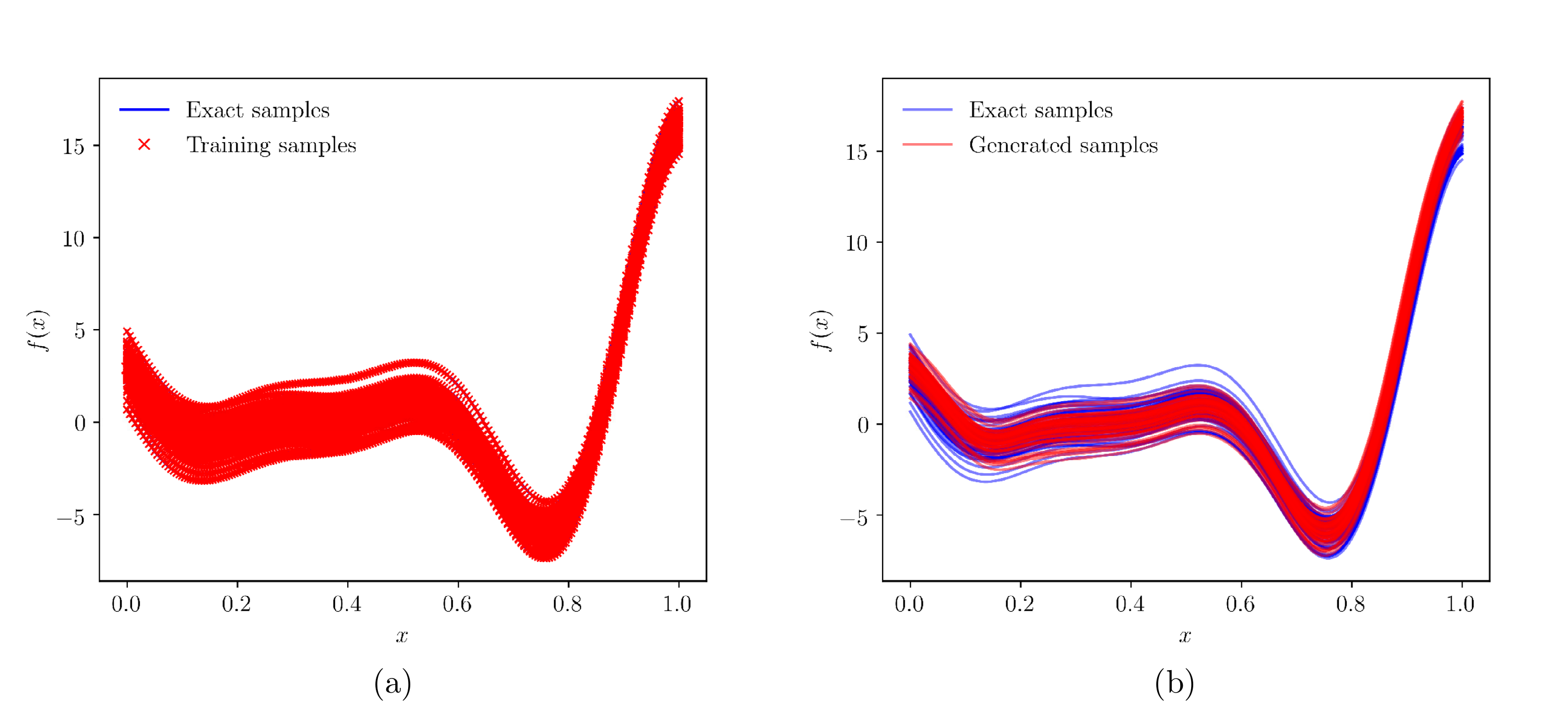}
\caption{{\em Sensitivity studies on approximating a one-dimensional stochastic process:} (a) Representative samples generated by this reference stochastic process, along with the observed data used for model training. (b) Representative samples generated by a conditional generative model with $\lambda=1.5$. Blue lines are the exact reference samples, red crosses are the training data, and red lines are the generated samples.}
\label{fig:app1}
\end{figure}

Here we provide results on a series of comprehensive systematic studies that aim to quantify the sensitivity of the resulting predictions on: 
\begin{enumerate}[label=(\roman*)]
    \item the entropic regularization penalty parameter $\lambda$.
    \item the generator, discriminator and encoder neural network architectures.
    \item the the adversarial training procedure.
\end{enumerate}
 To this end, we consider a simple benchmark corresponding to the approximation of a Gaussian process $g(x)\sim\mathcal{GP}(\mu_H(x), k(x,x';\theta_H))$, where $\mu_H(x)$ corresponds to the high-fidelity mean function defined in equation \ref{eq:mu_H} and $k(x,x';\theta_H)$ is a squared exponential kernel with hyper-parameters $\sigma_{f_H}^2 = 0.5, l_H^2=0.5$, as defined in equation \ref{eq:rbf_kernel}. Figure \ref{fig:app1}(a) shows representative samples generated by this reference stochastic process. In all cases we have employed simple feed-forward neural network architectures as described below. The comparison metric used in all sensitivity studies is the average discrepancy between the predicted and the exact one-dimensional marginal densities, as measured by the reverse Kullback-Leibler divergence
\begin{align}
    \mathbb{E}_{p(x)}\{\mathbb{KL}[p_1(y|x)||p_2(y|x)]\}
\end{align}
where $p_1(y|x)$ is the conditional distribution predicted by the generative model, $p_2(y|x)$ is the conditional distribution of the exact solution, and $p(x)$ is the distribution of uniformly sampled test locations  in the interval $x\in[0,1]$. For a given $x\sim p(x)$, we facilitate a tractable computation of the reverse KL-divergence using equation \ref{eq:KL_div}, by performing a Gaussian approximation of $p_2(y|x)$, while, by definition, $p_1(y|x)$ is a known uni-variate Gaussian density.

\subsection{Sensitivity with respect to the entropic regularization penalty parameter $\lambda$}\label{B2}

In this study we aim to quantify the sensitivity of our predictions with respect to the penalty parameter $\lambda$ in equation \ref{eq:generator_loss}. To this end, we have fixed the architecture for generator and encoder neural networks to include 3 hidden layers with 100 neurons each, and the discriminator neural network to include 2 hidden layers with 100 neurons each. In all cases, we have used a hyperbolic tangent non-linearity and a normal Xavier initialization \citep{glorot2010understanding}. For each iteration, we train the discriminator for 3 times and the generator for 1 time. We use a batch size of $500$ data-points per stochastic gradient update, and the total number of training points is $10000$. 

\begin{figure}
\centering
\includegraphics[width=\textwidth]{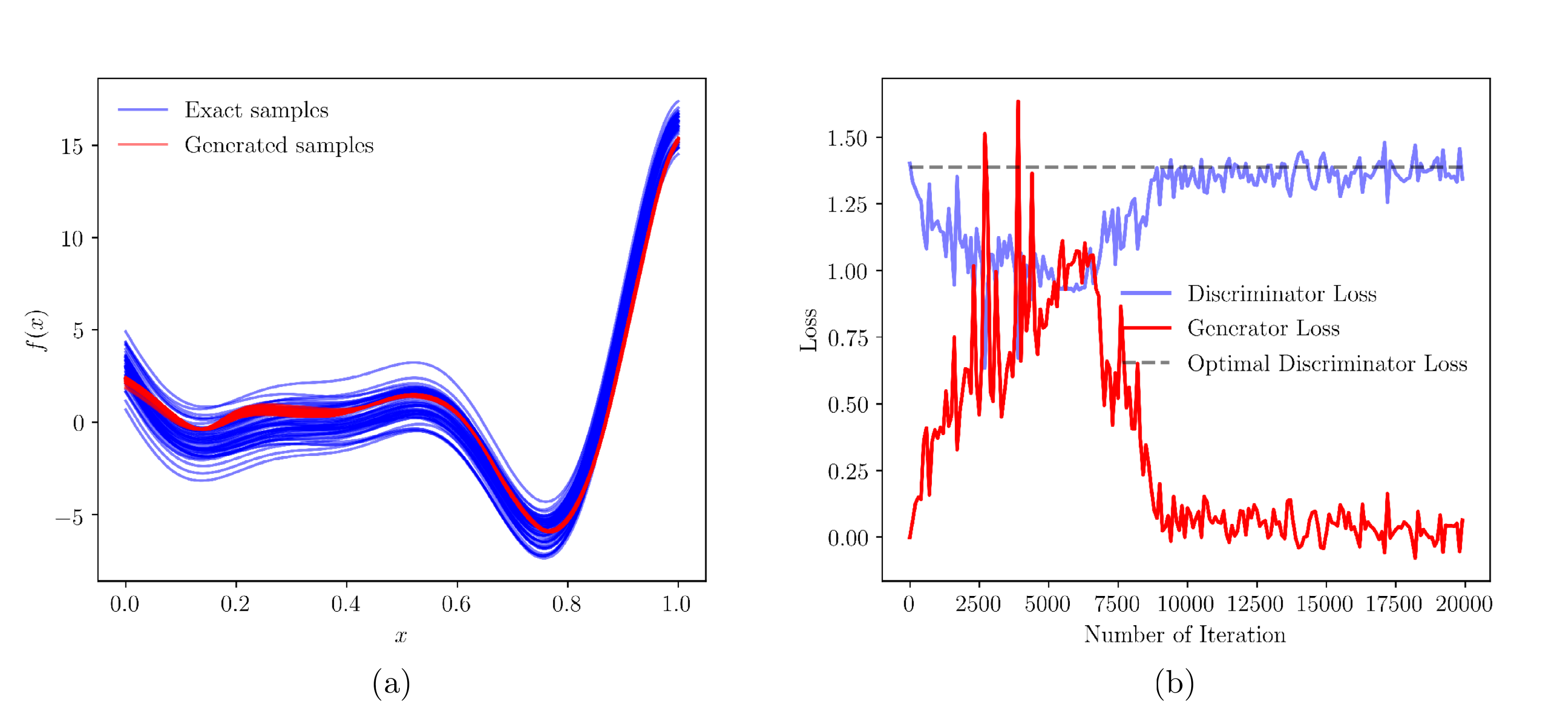}
\caption{{\em Sensitivity with respect to the entropic regularization penalty parameter $\lambda$:} (a) Manifestation of mode collapse for $\lambda=1.0$. Blue lines are exact samples from the reference stochastic process, red lines are samples produced by the conditional generative model. (b) Generator and discriminator loss values as a function of the number of training iterations.}
\label{fig:app_mode_collapse}
\end{figure}

In table \ref{tab:sens_t2} we report the reverse KL-divergence between the predicted data and the ground truth for different values of $\lambda$,  $1.0$, $1.2$, $1.5$, $1.8$, $2.0$, and $5.0$. Recall that for $\lambda=1.0$ our model has a direct correspondence with generative adversarial networks \cite{li2018learning}, while for $\lambda>1.0$ we obtain a regularized adversarial model that introduces flexibility in mitigating the issue of mode collapse. A manifestation of this pathology is evident in figure \ref{fig:app_mode_collapse}(a) in which the model with $\lambda=1.0$ collapses to a degenerate solution that severely underestimates the diversity observed in the true stochastic process samples, despite the fact that the model training dynamics seem to converge to a stable solution (see figure \ref{fig:app_mode_collapse}(b)). This is also confirmed by the computed average discrepancy in KL-divergence which is roughly an order of magnitude larger compared to the regularized models with $\lambda>1.0$. We also observe that model predictions remain robust for all values $\lambda>1.0$, while our best results are typically obtained for $\lambda=1.5$ which is the value used throughout this paper (see figure \ref{fig:app1}(b) for representative samples generated by the conditional generative model with $\lambda=1.5$).

\begin{table}[!htbp]
\centering
\begin{tabular}{|c|c|c|c|c|c|c|}
\hline
$\lambda$ & 1.0 & 1.2& 1.5& 1.8& 2.0& 5.0\\ 
\hline
Reverse-KL & 2.8e+00 & 2.2e-01& 2.2e-01& 4.2e-01 & 5.4e-01 & 3.4e-01 \\
\hline
\end{tabular}
\caption{{\em Sensitivity with respect to the entropic regularization penalty parameter $\lambda$:} Average reverse KL-divergence between the predicted and the ground truth one-dimensional marginals in $x\in[0,1]$ for different values of the entropic regularization penalty $\lambda$ in equation \ref{eq:generator_loss}.}
\label{tab:sens_t2}
\end{table}

\subsection{Sensitivity with respect to the neural network architecture}

In this study we aim to quantify the sensitivity of our predictions with respect to the architecture of the neural networks that parametrize the generator, the discriminator, and the encoder. Here, we choose the number of layers for the discriminator to always be one less than the number of layers for the generator and the encoder (e.g., if the number of layers for the generator is two then the number of layers for the discriminator is one, etc.). In all cases, we fix $\lambda = 1.5$ and we use a hyperbolic tangent non-linearity, and a normal Xavier initialization \citep{glorot2010understanding}. In table \ref{tab:sens_t3} we report the computed average reverse KL-divergence between the predicted data and the ground truth for different feed-forward architectures for the generator, discriminator, and encoder (i.e., different number of layers and number of nodes in each layer). We denote the number of neurons in each layer as $N_{n}$ and the number of layers for the  generator and the encoder as $N_g$. 

The results of this sensitivity study are summarized in table \ref{tab:sens_t3}. Overall, we observe that model predictions remain robust for all neural network architectures considered.

\begin{table}[!htbp]
\centering
\begin{tabular}{|c|c|c|c|}
\hline
\diagbox{$N_g$}{$N_n$} & 20& 50& 100\\ 
\hline
2 &    6.0e-01& 6.0e-01& 7.4e-01 \\
\hline
3 &    9.6e-02& 3.3e-01& 2.2e-01 \\
\hline
4 &    1.5e-01& 4.0e-01& 2.7e-01 \\
\hline
\end{tabular}
\caption{{\em Sensitivity with respect to the neural network architecture:} Average reverse KL-divergence between the predicted data and the ground truth for different feed-forward architectures for the generator, encoder, and the discriminator. The total number of layers of the latter is always chosen to be one less than the number of layers for generator.}
\label{tab:sens_t3}
\end{table}


\subsection{Sensitivity with respect to the adversarial training procedure}

As discussed in \cite{yang2018adversarial}, the adversarial training procedure plays a key role in the effectiveness of adversarial generative models, and it often requires a careful tuning of the training dynamics to ensure robustness in the model predictions. To this end, here we test the sensitivity of the proposed conditional generative model with respect to the relative frequency in which the generator and discriminator networks are updated during model training. To this end, we fix we the entropic regularization penalty to $\lambda = 1.5$, use the neural network architecture to be the same as the one described in section \ref{B2}, and vary the total number of training steps for the generator $K_g$ and the discriminator $K_d$ within each stochastic gradient descent iteration. 

The results of this study are presented in table \ref{tab:sens_t4} where we report the average reverse KL-divergence between the predicted data and the ground truth. These results reveal the high sensitivity of the training dynamics on the interplay between the generator and discriminator networks, and pinpoint the well known peculiarity of adversarial inference procedures which require a careful tuning of $K_g$ and $K_d$ for achieving stable performance in practice. Overall we observe that a one-to-three or one-to-five ratio of relative updates for the generator and discriminator, respectively, is the setting that typically works best in practice, although we must underline that this also depends on the capacity of the underlying neural network architectures as discussed in  \cite{yang2018adversarial}.

\begin{figure}
\centering
\includegraphics[width=0.6\textwidth]{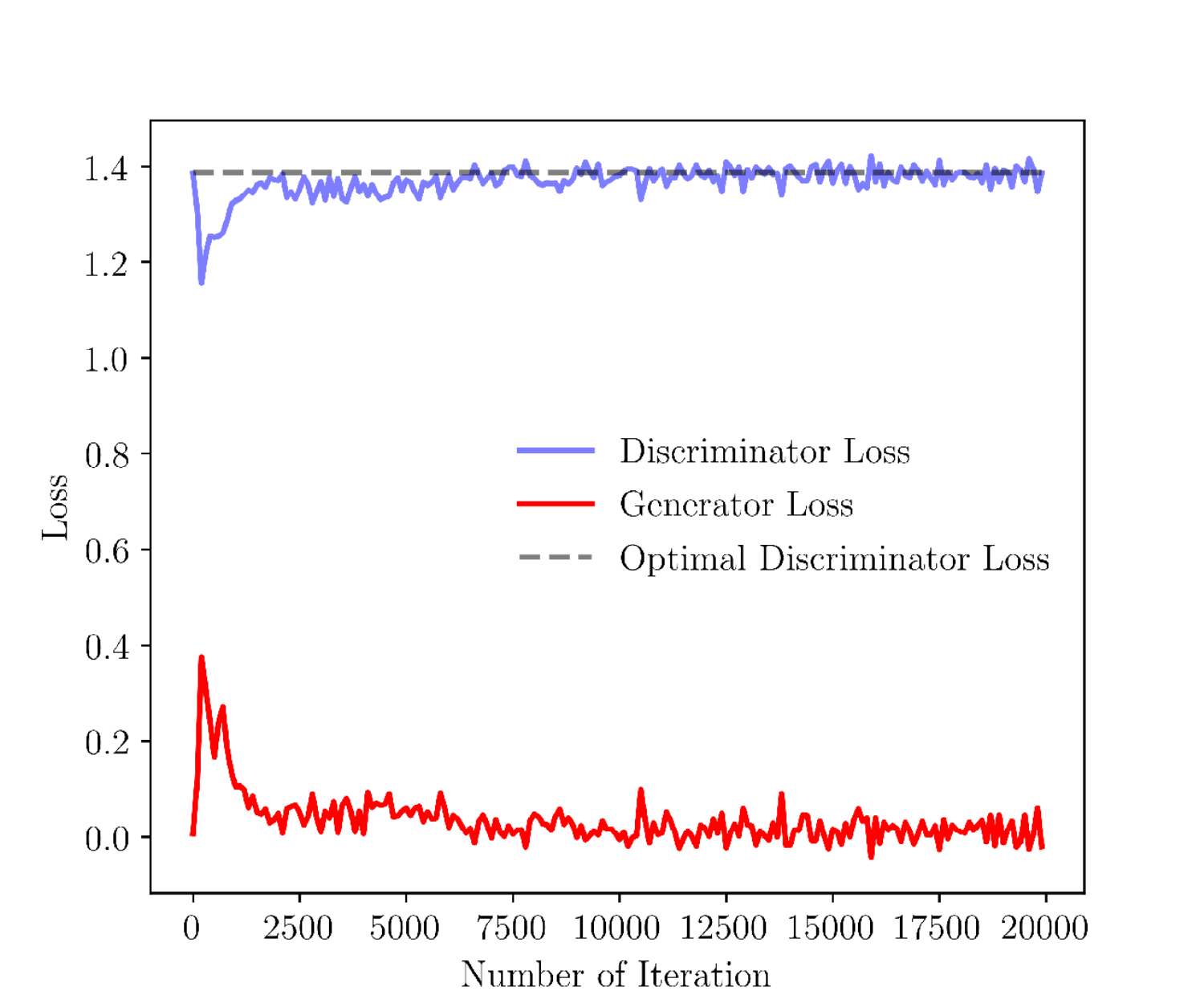}
\caption{{\em Optimal convergence of the discriminator loss:}  Convergence of the training algorithm for the case $K_g=1$ and $K_d=5$. The red line depicts the generator loss, the blue line is the discriminator loss, and the black dash line is the theoretical optimal loss of the discriminator.}
\label{fig:loss}
\end{figure}

Finally, figure \ref{fig:loss} depicts the convergence of the training algorithm for the case $K_g=1$ and $K_d=5$. According to \cite{goodfellow2014generative}, the theoretical optimal value of the discriminator loss is $\ln(4) = - 2 \times \ln(0.5) = 1.384$. As is shown in figure \ref{fig:loss}, the losses oscillate at the very beginning of the training and quickly converge to the optimal value after approximately 2,000 iterations. 

\begin{table}[!htbp]
\centering
\begin{tabular}{|c|c|c|c|}
\hline
\diagbox{$K_g$}{$K_d$} & 1& 3& 5\\ 
\hline
1 &    9.2e-01& 2.2e-01& 2.4e-01 \\
\hline
3 &    1.2e+00& 8.5e-01& 9.4e-01 \\
\hline
5 &    4.3e+00& 8.9e-01& 5.9e+00 \\
\hline
\end{tabular}
\caption{{\em Sensitivity with respect to the adversarial training procedure:} Average reverse KL-divergence between the predicted data and the ground truth with different number of relative updates between the generator and discriminator in each stochastic gradient descent iteration.}
\label{tab:sens_t4}
\end{table}

\end{document}